\documentclass[letterpaper]{article} 
\usepackage[]{aaai24}  
\usepackage{times}  
\usepackage{helvet}  
\usepackage{courier}  
\usepackage[hyphens]{url}  
\usepackage{graphicx} 
\usepackage{natbib}  
\usepackage{caption} 
\usepackage{algorithm}
\usepackage{algorithmic}
\usepackage{xcolor}
\usepackage{dsfont}
\usepackage{subcaption}
\usepackage{amsmath}
\usepackage{amssymb}
\usepackage{multirow}
\usepackage{ctable}

\usepackage{dblfloatfix}
\usepackage{bbding}

\usepackage{color}
\definecolor{Red}{cmyk}{0,1,1,0}
\definecolor{Green}{cmyk}{1,0,1,0}
\definecolor{Cyan}{cmyk}{1,0,0,0}
\definecolor{Purple}{cmyk}{0.45,0.86,0,0}
\definecolor{Rosolic}{cmyk}{0.00,1.00,0.50,0}
\definecolor{Blue}{cmyk}{1.00,1.00,0.00,0}
\definecolor{Orange}{cmyk}{0,0.52,0.80,0}
\definecolor{Black}{cmyk}{1,0,0,1}

\usepackage{newfloat}
\usepackage{listings}
\DeclareCaptionStyle{ruled}{labelfont=normalfont,labelsep=colon,strut=off} 
\floatstyle{ruled}
\newfloat{listing}{tb}{lst}{}
\floatname{listing}{Listing}
\title{O$^2$-Recon: Completing 3D Reconstruction of Occluded Objects in the Scene \\
with a Pre-trained 2D Diffusion Model}

\author{Yubin Hu\textsuperscript{\rm 1}, Sheng Ye\textsuperscript{\rm 1}, Wang Zhao\textsuperscript{\rm 1}, Matthieu Lin\textsuperscript{\rm 1}, Yuze He\textsuperscript{\rm 1}, \\ Yu-Hui Wen\textsuperscript{\rm 3}, Ying He\textsuperscript{\rm 2}, Yong-Jin Liu\textsuperscript{\rm 1}}

\affiliations{
\textsuperscript{\rm 1}Tsinghua University \\
\textsuperscript{\rm 2}Nanyang Technological University \\ 
\textsuperscript{\rm 3}Beijing Jiaotong University \\
}

\usepackage{bibentry}

\begin{document}

\maketitle

\begin{abstract}
Occlusion is a common issue in 3D reconstruction from RGB-D videos, often blocking the complete reconstruction of objects and presenting an ongoing problem. In this paper, we propose a novel framework, empowered by a 2D diffusion-based in-painting model, to reconstruct complete surfaces for the hidden parts of objects.
Specifically, we utilize a pre-trained diffusion model to fill in the hidden areas of 2D images. Then we use these in-painted images to optimize a neural implicit surface representation for each instance for 3D reconstruction.
Since creating the in-painting masks needed for this process is tricky, we adopt a human-in-the-loop strategy that involves very little human engagement to generate high-quality masks.
Moreover, some parts of objects can be totally hidden because the videos are usually shot from limited perspectives. To ensure recovering these invisible areas, we develop a cascaded network architecture for predicting signed distance field, making use of different frequency bands of positional encoding and maintaining overall smoothness.
Besides the commonly used rendering loss, Eikonal loss, and silhouette loss, we adopt a CLIP-based semantic consistency loss to guide the surface from unseen camera angles. 
Experiments on ScanNet scenes show that our proposed framework achieves state-of-the-art accuracy and completeness in object-level reconstruction from scene-level RGB-D videos. Code: \url{https://github.com/THU-LYJ-Lab/O2-Recon}.
\end{abstract}

\section{Introduction}

The task of reconstructing 3D objects within a scene has been a longstanding challenge in computer vision. Unlike scene-level reconstruction techniques~\cite{neural-rgbd,go-surf}, object-level 3D reconstruction focuses on creating individual representations for each instance within a scene. This technique is crucial for applications in computer vision, robotics, and mixed reality that require fined-grained scene modeling and understanding. 



Many works approach object-level 3D reconstruction as a task of estimating an object's pose and shape code, using a categorical generative model \cite{frodo,ellip-sdf}. 
While these methods create complete shapes, they are limited to reconstructing objects from specific categories, like tables or chairs. 
Even within these categories, the generated shape codes often struggle to accurately match the actual object surfaces. 
There are also a few approaches focusing retrieving suitable CAD models from a database and estimating their 9 degrees of freedom poses \cite{scan2cad}. These methods also face similar issues, such as limited scalability and low accuracy in reconstruction. 

\begin{figure}[tb]
  \centering 
  \includegraphics[width=0.9\columnwidth]{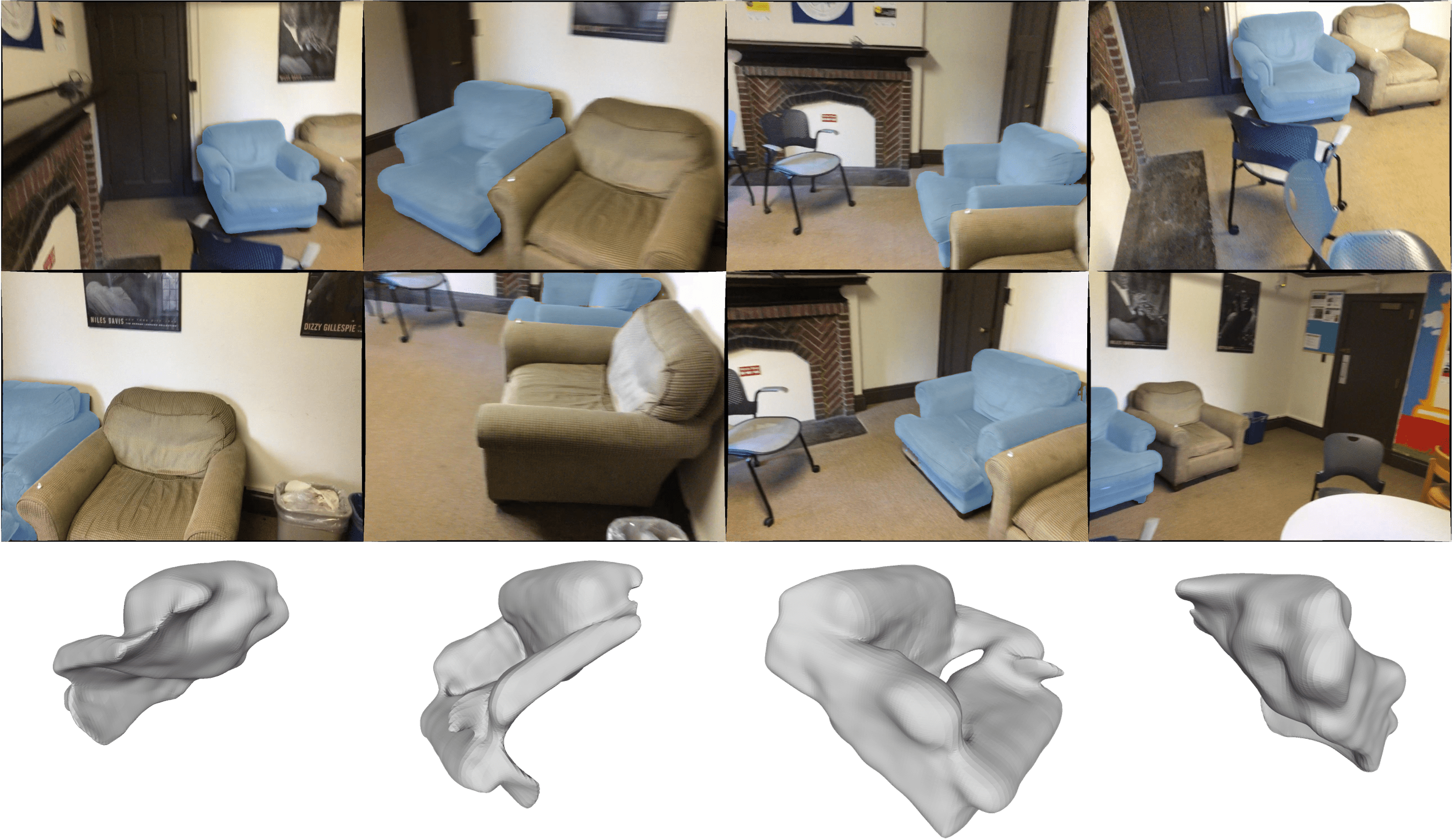}
  \vspace{-2mm}
  \caption{%
    Occlusion presents significant hurdles for object-level reconstruction. For the occluded armchair, highlighted in blue, existing methods can only yield a partial reconstruction where a significant portion of the geometry is missing.
  }
  \label{fig:intro_demo}
  \vspace{-2mm}
\end{figure}

Benefiting from the emerging technology of neural radiance fields (NeRF) \cite{nerf,imap}, vMap \cite{vmap} is able to reconstruct a wider variety of objects, moving beyond just categorical instances. 
However, it does not address the issue of occlusion in scene-level videos, which results in incomplete observations of objects and reduced reconstruction quality. 
As illustrated in Figure \ref{fig:intro_demo}, the camera paths in 3D indoor scenes often limit the coverage of scene-level videos. 
As a result, objects close to walls or to each other are frequently only partially recorded. 
The lack of complete visuals, especially the absence of information for the occluded regions, makes these images inadequate for neural rendering-based reconstruction methods \cite{neus, volsdf}.

Inspired by the recent success of diffusion-based image in-painting \cite{Wang_2023_CVPR}, we explore the application of a pre-trained diffusion model to in-paint the occluded regions in the input video frames. 
While the latent diffusion model \cite{stable-diffusion} is adept at in-painting missing regions in images,
it may produce drastically incorrect content without precise in-painting masks that identify the missing parts.
In this paper, we address this challenge by introducing affordable human interaction into our framework, thereby ensuring both the accuracy of the masks and the overall quality of the in-painting process.

Provided with an RGB-D video sequence accompanied by object masks, our system requires a user to choose between 1 to 3 frames containing occlusion. The user is then guided to sketch the in-painting masks for these frames, utilizing their experience and judgement.
These sketched masks are subsequently re-projected to all other views, utilizing depth information in-painted by the diffusion model, and then merged to create the in-painting masks for the remaining frames.
By incorporating cost-effective human engagement, our proposed approach ensures the generation of high-quality in-painting masks. These masks maintain robust geometric consistency across various views, thereby guiding the 2D diffusion model to create  convincing and coherent in-paintings for the occluded regions.
As for the reconstruction stage, we utilize the neural implicit surface representation like NeuS \cite{neus} and optimize it with rendering loss.
Given the possible visual inconsistency across the in-painted images, the implicit representation can filter the inconsistency during the multi-view rendering-based optimization and reconstruct reasonable underlying surfaces.

\begin{figure*}[tb]
  \centering 
  \includegraphics[width=0.95\linewidth]{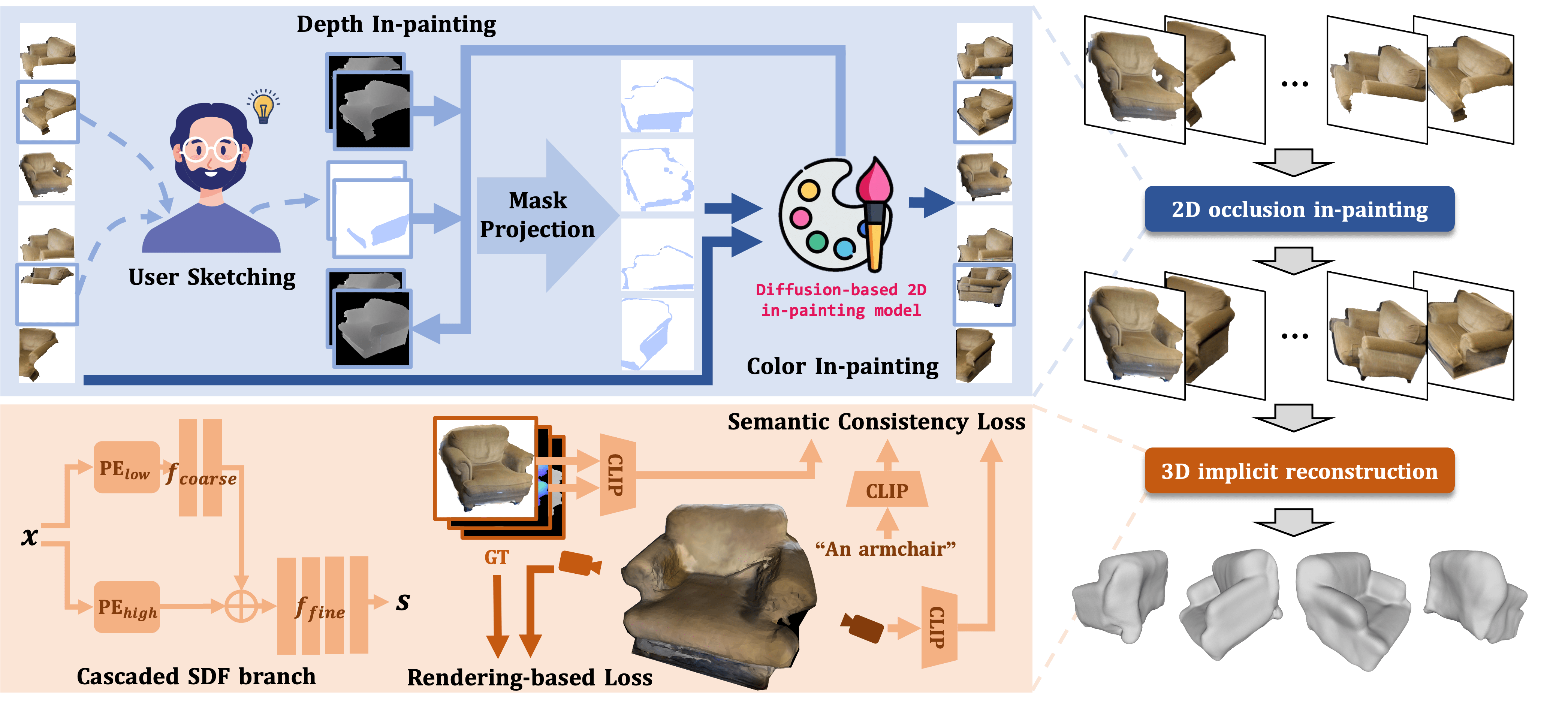}
  \vspace{-3mm}
  \caption{%
    The proposed O$^2$-Recon framework. We utilize the Stable Diffusion in-painting model\protect\footnotemark \ in our implementation.
  }
  \label{fig:main}
\end{figure*}

To mitigate the reconstructed artifacts in areas that are entirely unseen, our system enhances the rendering-based reconstruction from two perspectives: first, by adopting  semantic supervision over the unseen regions; second, by applying a smoothness prior of the neural implicit surface. In the case of semantic supervision, we guide the reconstruction by supervising the CLIP \cite{clip} features of renderings from novel views within both the image and text domains. For smoothness, we introduce a cascaded architecture for predicting signed distance field (SDF), which is specially designed to prevent noisy artifacts in the unseen regions. To achieve this, we utilize a shallow MLP equipped with low-frequency positional encodings (PEs), ensuring overall smoothness of the surface. Concurrently, we adopt a deeper auxiliary branch, armed with high-frequency PEs, to predict residuals of SDF. This dual approach is effective in maintaining superior expressiveness of visible regions while ensuring a balanced and coherent reconstruction.

To sum up, the main contributions of our work include: 

1) A 3D reconstruction framework for occluded objects in the scene, termed O$^2$-Recon, that addresses the occlusion problem by employing diffusion-based in-painting within the 2D image domain. 2) A human-in-the-loop strategy for in-painting mask generation, enabling the production of high-quality masks with minimal human engagement, which are used to guide the diffusion-based 2D in-painting process. 3) The creation of a novel cascaded SDF prediction network, coupled with semantic consistency supervision using CLIP, to enhance the surface quality of completely unseen regions in occluded objects. 



We conduct extensive experiments on ScanNet scenes, demonstrating that our proposed framework achieves state-of-the-art reconstruction accuracy and completeness for occluded objects in the scene.  
With the complete objects reconstructed by our method, we enable further object-level manipulations with highly free translations and rotations.

\section{Related Works}

\textbf{Object-Level 3D Scene Reconstruction.} 
Generating an independent 3D representation for individual objects within a scene is an active research area. Many methods seek to address this problem through the joint optimization of an object's shape code and pose. For instance, 
FroDO \cite{frodo} utilizes a pre-trained encoder-decoder network inspired by DeepSDF \cite{deepsdf} to map RGB images to a sparse point cloud and a dense SDF field using a latent shape code as a proxy. 
ELLIPSDF \cite{ellip-sdf} introduces a bi-level object model 
that captures both the coarse-level scale and the fine-level shape details, enhancing the joint optimization process for object pose and shape code. 
To enable real-time reconstruction, MOLTR \cite{moltr} removes the backward optimization and focuses on predicting the shape code by multi-view image encodings.
It leverages another pre-trained 3D detector to predict the objects' 9-DoF poses. 
Departing from the typical two-stage pipeline, CenterSnap \cite{centersnap} and RayTran \cite{raytran}  unifies pose and shape estimation into a single-stage network.
Instead of predicting shape codes, RayTran directly predicts the SDF volume.
Despite their ability for reconstructing complete shapes for individual objects, these methods are typically constrained to specific categories such as tables or chairs.
Additionally, models that are pre-trained on synthetic datasets like ShapeNet \cite{shapenet} often struggle when applied to real-world scenarios, since the surfaces decoded from shape codes might not accurately represent actual objects. 

Another class of methods leverages CAD databases instead of the generative models, retrieving suitable models \cite{scan2cad} and applying deformations \cite{cad-deform} to align with actual objects. However, the inherent limitation of deformation operation implies that these methods lack the flexibility to accurately represent real-world objects and the ability for high-fidelity reconstruction.

There are also approaches that utilize NeRF~\cite{nerf} for object-level reconstruction of arbitrary 3D objects. For example, vMap \cite{vmap} represents each object with an independent NeRF, and optimizes it through photometric loss. While effective in certain scenarios, this approach fails to handle occlusion, often resulting in incomplete and degenerated surfaces when parts of objects are not visible. 
Object-NeRF \cite{objectnerf} addresses the misleading supervision of incomplete instance masks with the use of a 3D guard mask, however it still relies on the intrinsic smoothness bias of NeRF (similar to vMap) to mitigate the occluded regions.
RICO \cite{rico} regularizes the unseen areas through object-background relationship,  but it falls short in providing effective supervision for the occluded parts, leaving room for further improvement. 

Our approach differs from the above methods in that it explicitly supervises the occluded regions using in-paintings generated by a pre-trained 2D diffusion model. It offers two unique features. Firstly, it reconstructs accurate surfaces for \textit{arbitrary} objects by relying on a neural implicit surface representation. Secondly, the application of diffusion-based in-painting model enables our method to reconstruct \textit{complete} shapes of objects, even when they are partially occluded.

\noindent\textbf{NeRFs Empowered by 2D Diffusion Models. } 
The success of diffusion models in 2D image generation and editing \cite{imagen, imagic} has motivated interest in combining these advanced 2D models with NeRF representations. 
To achieve NeRF-level editing guided by text instructions, instruct-NeRF2NeRF \cite{Instruct-NeRF2NeRF} iteratively updates the multi-view image dataset with the edited images, which are rendered by the pre-trained InstructPix2Pix model \cite{instruct-pix2pix}. 
Guided by the pre-trained Stable Diffusion model, the recently proposed RePaint-NeRF \cite{RePaint-NeRF} facilitates local editing within selected areas in NeRF scenes.
There are also works for generating NeRFs from text prompts with the aid of 2D diffusion models by different approaches, such as score distillation sampling \cite{dreamfusion, magic3d}, score Jacobian chaining \cite{scoreJacobian}, and variational score distillation \cite{ProlificDreamer}.

In this paper, we utilize a diffusion-based 2D in-painting model to aid 3D reconstruction of occluded objects. 
The pre-trained diffusion model is used to in-paint the occluded regions in the 2D images. These enhanced images subsequently serve as the foundation to reconstruct complete shapes for occluded objects.

\section{Method}

\footnotetext{https://huggingface.co/runwayml/stable-diffusion-inpainting}

Given an RGB-D video clip composed of $N$ image frames $\{I_n\}_{n=1}^{N}$ and depth frames $\{D_n\}_{n=1}^{N}$, we assume that high-quality instance and semantic segmentation results $\{S^I_n\}_{n=1}^{N}$ and $\{S^S_n\}_{n=1}^{N}$ are already obtained by existing methods such as  \cite{vmap,segformer}. 
In this paper, we aim to reconstruct the complete shapes of occluded objects.
As shown in Figure \ref{fig:main}, our method begins by in-painting the occluded regions in images, utilizing a pre-trained diffusion model, which in our implementation is the Stable Diffusion in-painting model \cite{stable-diffusion}. 
We then reconstruct the 3D object using a neural implicit surface representation that compensates for the entirely unseen regions (an example is provided in the rightmost part of Figure \ref{fig:main}).

In Section \ref{sec:2d_inpaint}, we elaborate on our proposed 2D in-painting process for occluded objects in images, employing the pre-trained diffusion model with minimal human engagement. In the subsequent neural implicit surface based reconstruction, we design a cascaded network architecture for the SDF branch, effectively preventing degenerated high-frequency artifacts in the unseen areas (see Section \ref{sec:cascaded_network}). Finally, we discuss the loss functions utilized in the entire optimization process in Section \ref{sec:loss_function}.

\subsection{Diffusion-based 2D Occlusion In-painting}
\label{sec:2d_inpaint}

Utilizing the instance segmentation, we first extract the object mask $\{M_n^i\}_{n=1}^{N}$ for each object with the identifier $i$: 
\begin{equation}
    M_n^i = \mathds{1}_{A_i}(x,y), \quad A_i=\{(x,y) | S_n^I(x,y)=i \}.
\end{equation}
Subsequently, we apply the Hadamard product to the extracted masks and RGB-D frames. This process yields the masked RGB images $\{I_n^i\}_{n=1}^{N}$ and depths $\{D_n^i\}_{n=1}^{N}$ for each object $i$, as defined by
\begin{equation}
    I_n^i = I_n \circ M_n^i, \quad D_n^i = D_n \circ M_n^i.
\end{equation}
These masked data, typically incomplete for occluded objects, can present incorrect boundaries that may disrupt downstream rendering-based geometry optimization.
We address this challenge by completing the occluded objects in images using a pre-trained diffusion model.

\begin{figure}[t]
\centering

     \begin{subfigure}[b]{0.3\linewidth}
         \centering
         \includegraphics[width=1.0\linewidth]{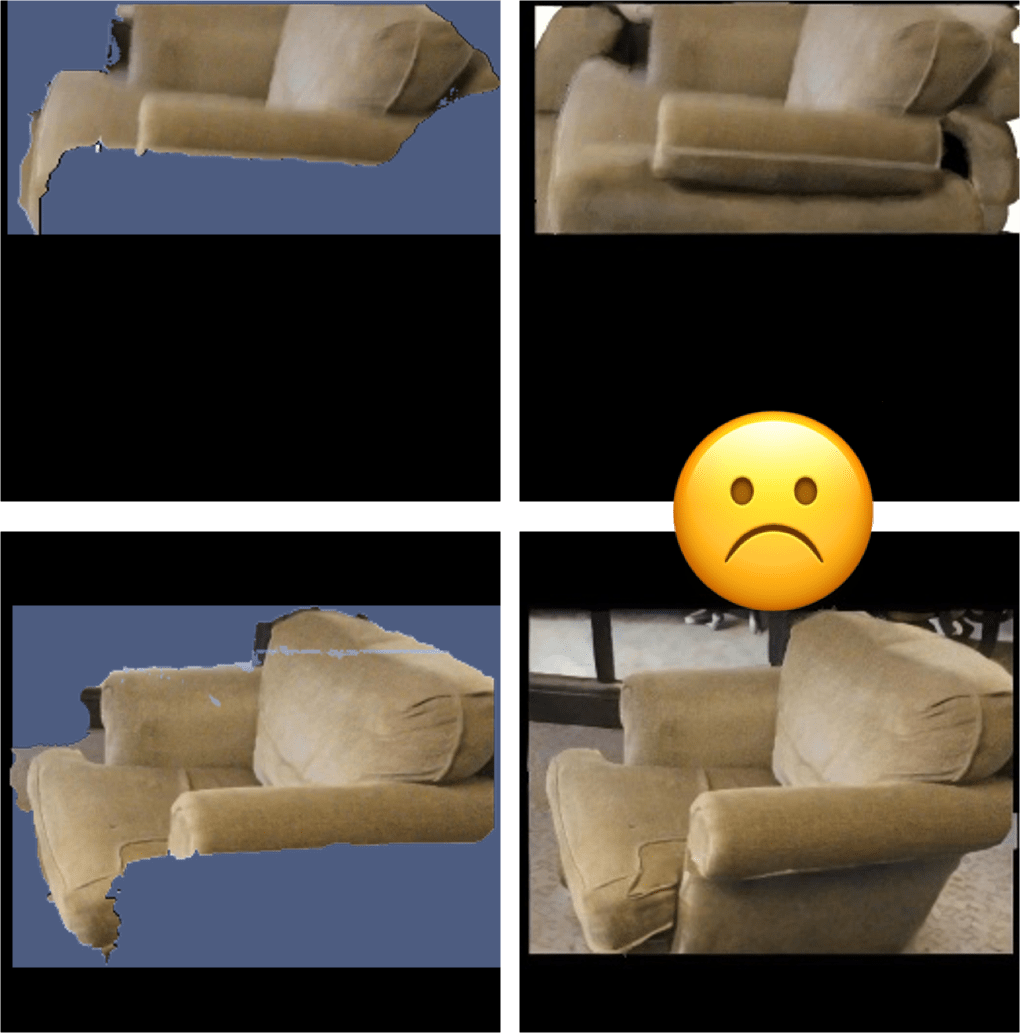}
         \caption{}
         \label{fig:box-mask}
     \end{subfigure}
     \hfill
     \begin{subfigure}[b]{0.3\linewidth}
         \centering
         \includegraphics[width=1.0\linewidth]{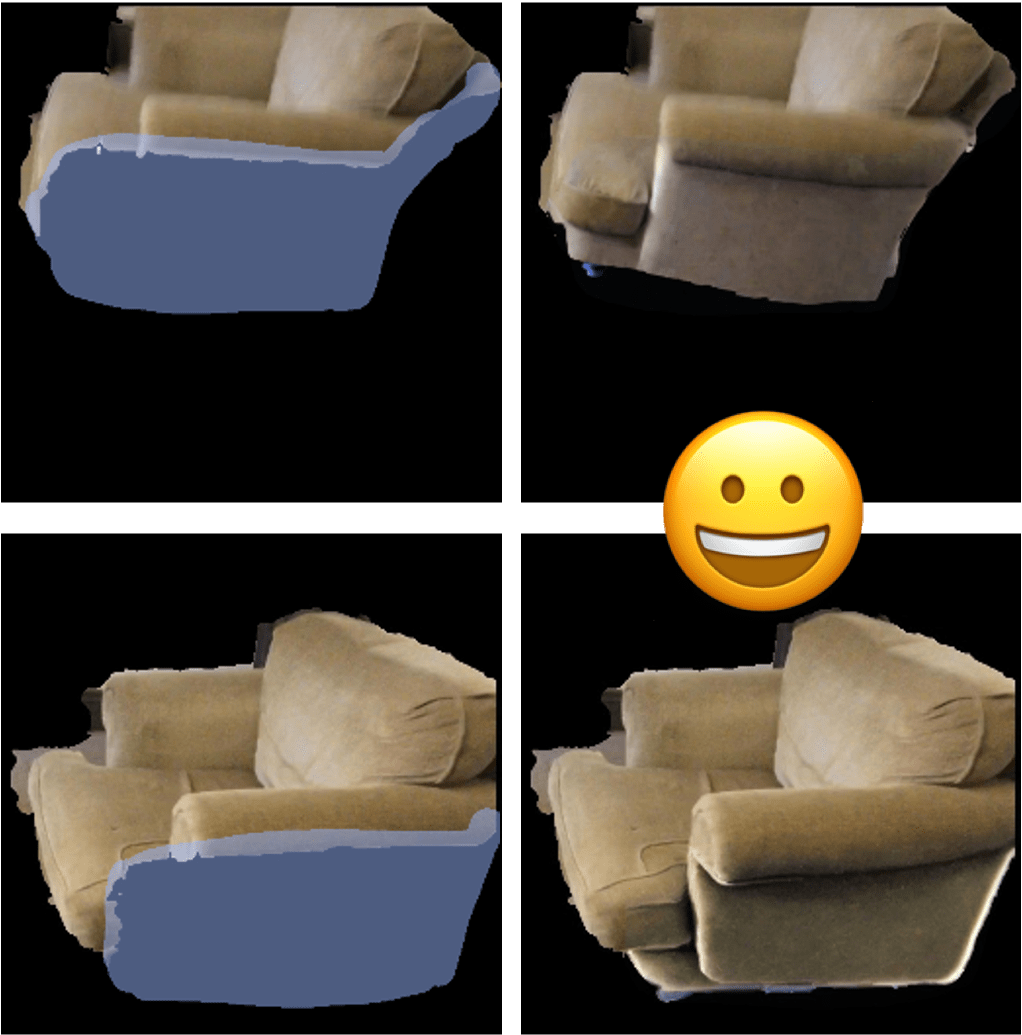}
         \caption{}
         \label{fig:draw-mask}
     \end{subfigure}
     \hfill
     \begin{subfigure}[b]{0.3\linewidth}
         \centering
         \includegraphics[width=1.0\linewidth]{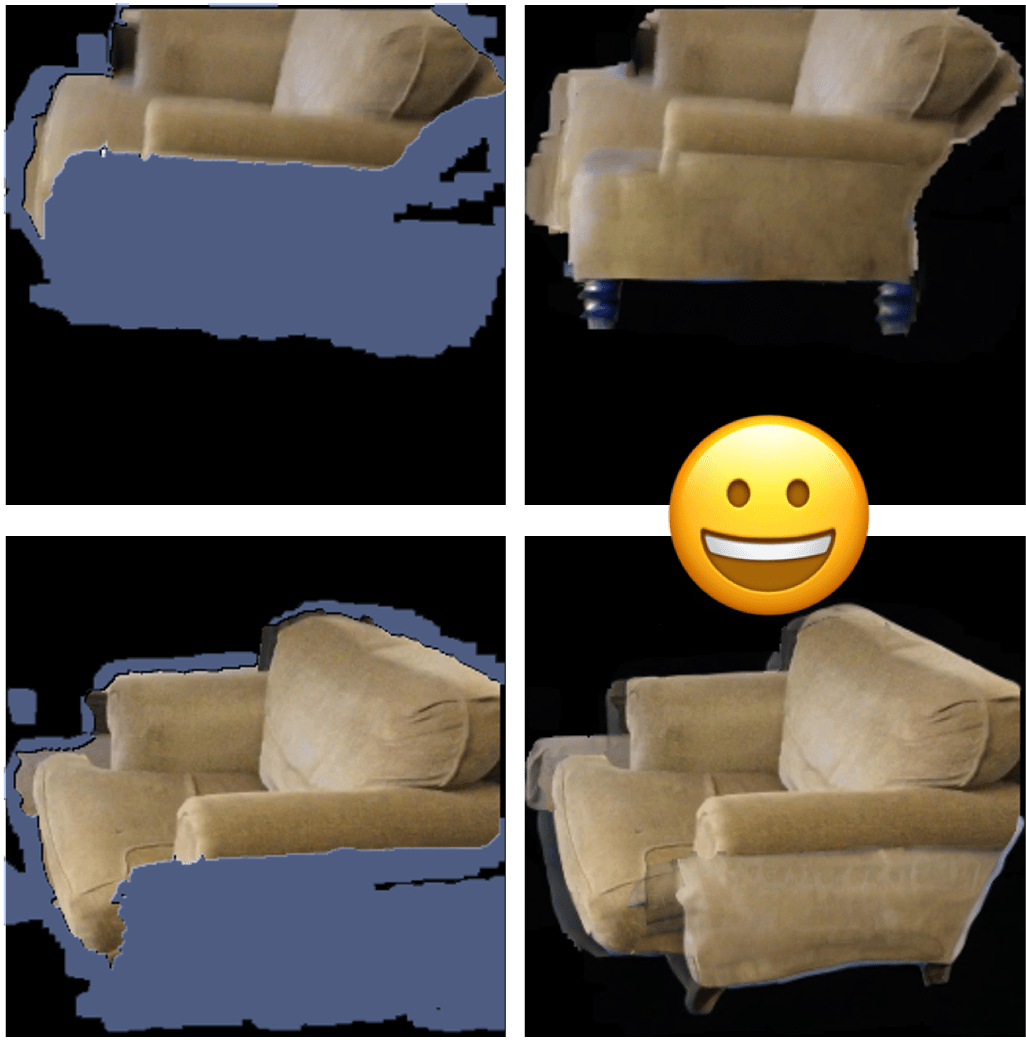}
         \caption{}
         \label{fig:project-mask}
     \end{subfigure}
     
\vspace{-3mm}
\caption{Illustration of results produced by different in-painting masks: (a) bounding box masks, (b) user-sketched masks, and (c) masks projected from selected views.
}
\label{fig:mask-prompt}

\end{figure}

Note that to achieve satisfactory  in-painting results, text prompts and high-quality mask prompts must be provided to the 2D diffusion model. However, generating accurate in-panting masks for occluded objects is a highly non-trivial task. 
Simply utilizing the 2D bounding box of the visible region as an in-painting mask may lead to background contents appearing inside the object region. Sometimes, the bounding box only encompasses part of the whole object, resulting in incomplete in-painting, as shown in Figure \ref{fig:box-mask}. 
Moreover, since the occluded areas may vary significantly between different views, predicting geometrically consistent in-painting masks through automated algorithms poses a technical challenge.
To overcome the challenge, we propose a human-in-the-loop mask generation strategy that requires minimal human engagement.

\textbf{User Sketching.} 
As shown in the upper part of Figure \ref{fig:main}, we enlist the assistance of a user to sketch the in-painting mask on 1 to 3 representative images. This process does not necessitate specialized expertise from the participants and can be completed in just 1 to 2 minutes for each object. 

\textbf{Depth In-painting.} Building upon the user-sketched 2D masks, we aim to re-project these masks to generate the in-painting masks for all other frames.
However, this operation is complicated by the absence of depth information in the sketched area due to occlusion. To project in-painting masks from the sketched images to the correct regions of other images, we utilize the pre-trained diffusion model to predict pseudo depth for the sketched area. By treating the depth map as a grayscale image, we feed both the masked depth frame and the sketched in-painting masks into the diffusion model, which yields a predicted completed depth map.

\textbf{Mask Projection.} We formulate the mask projection as: 
\begin{equation}
    \tilde{M}_{m}^i = Merge(\{\tilde{M}_{n\to m}^i | n \in \text{selected \ views}\}),
\end{equation}
\begin{equation}
    \tilde{M}_{n\to m}^i = Proj(\tilde{M}_{n}^i, P_n, P_m, K_m),
\end{equation}
where $n$ is the source view (i.e., the view containing user sketches) and $m$ denotes the target view. $\tilde{M}_{m}^i$ represents the in-painting mask for object $i$ in  frame $m$. $P_m$ and $K_m$ are the extrinsic and intrinsic matrix of the depth frame $m$. $Proj(\cdot)$ and $Merge(\cdot)$ denote the mask projection and merging process, respectively.  

\textbf{Color In-painting.} We take the in-painting masks and incomplete color images as inputs and feed them into the diffusion model for in-painting. Thanks to the human-in-the-loop strategy, we are able to generate high-quality in-painting masks $\{\tilde{M}_{n}^i\}_{n=1}^N$, which effectively guide and enhance the filling process.
Though these projected masks might not be precisely accurate, they serve as valuable indicators of the visible contours for the occluded objects. This helps to prevent the intrusion of background contents into the object region and effectively directs the creation of plausible object shapes, as illustrated in Figure \ref{fig:project-mask}.

\textbf{Mask Refining.} Once the in-painting has been completed, we further refine the object masks according to the in-painted RGB images. 
The updated masks for object $i$ are denoted by $\{\hat{M}_{n}^i\}_{n=1}^N$ .
For both depth and color in-painting processes, we adopt text prompts that reflect the semantic class identified in $\{S^S_n\}_{n=1}^{N}$ by \textit{``A/an \$\{CLASS\}."}
We refer readers to the supplementary material for more details about mask projection and refining. 

\subsection{Cascaded SDF Prediction}
\label{sec:cascaded_network}

In scene-level videos, the limited number of camera views available for each individual object falls short in guiding rendering-based optimization, creating a unique challenge in sparse-view object reconstruction. As highlighted in recent works \cite{sparse_prior, sparseneus, volrecon}, the SDF network involved in neural implicit surface reconstruction tends to overfit to the color appearance, rather than accurately learning the surface geometry. This overfitting often leads to artifacts, such as degeneration in the unseen regions.

To tackle this issue, we propose a cascaded network architecture to enhance the smoothness priors in the SDF branch.
As shown in the bottom-left section of Figure \ref{fig:main}, our architecture adopts a two-part structure, differing from popular neural implicit surfaces, such as NeuS \cite{neus} that typically uses a single large MLP. The first part is a coarse prediction block with low-frequency positional encodings $PE_{low}$ and shallow MLP layers $f_{coarse}$. The second part is a refinement block, employing high-frequency positional encodings $PE_{high}$ and deep MLP layers $f_{fine}$.

\begin{figure*}[!hbp]
  \centering 
  \includegraphics[width=0.95\linewidth]{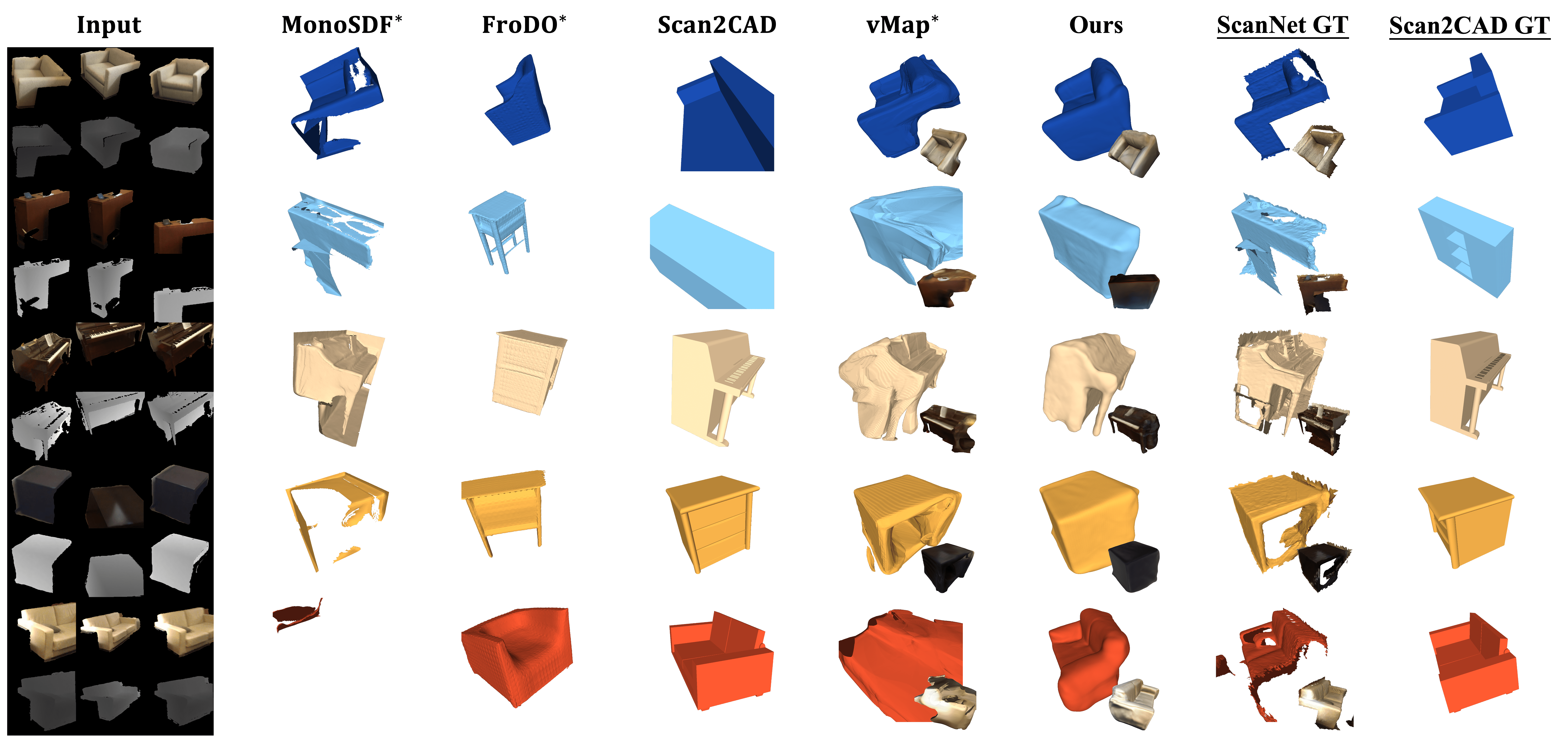}
  \caption{%
    Visualization of occluded objects reconstructed by different methods. The occlusion conditions can be visualized in the column of Input and ScanNet GT, the missing parts indicate occlusion in the corresponding regions.
  }
  \label{fig:main-exp}
\end{figure*}

We train the cascaded network using a two-stage strategy that separates low-frequency geometry and high-frequency fine details.
In the first stage, we focus on training the coarse SDF prediction block for generating a smooth surface. This initial stage is pivotal as it guards against rapid overfitting to the restricted camera views, thereby providing a stable initialization. Recognizing surfaces obtained in the first stage may be over-smoothed and lack fine details, the second stage comes into play. In this phase, we activate the refinement block, working in conjunction with the coarse block, to predict SDF residuals. 
In the cascaded network, the SDF value $s$ at a certain 3D point $x$ is formulated as
\begin{equation}
    s = f_{fine}([f_{coarse}(PE_{low}(x)), PE_{high}(x)]).
\end{equation}

Experiments show that our two-stage training strategy, separating the learning of low-frequency and high-frequency signals,  stabilizes the training process while enhancing the network's ability of capturing fine-grained geometry of visible areas. This approach strikes a balance between overall smoothness and intricate detail. Furthermore, by employing this cascaded architecture, we enhance the reconstruction quality of entirely unseen surfaces, enabling the manipulation of reconstructed objects even under large rotations.


\subsection{Loss Functions}
\label{sec:loss_function}

Given the in-painted images, the updated object masks, and the original depth information, we optimize the implicit representation with the sum of following loss functions.

\textbf{Rendering-based Loss.} Using the volume rendering equation in MonoSDF \cite{monosdf}, we can render the expected color, normal and depth values of ray $\textbf{r}$ and supervise them with the ground truth values. 
The ground truth surface normal maps are predicted from the in-painted RGB images by SNU \cite{snu} following the practice in NeuRIS \cite{neuris}. 
For the color and normal values, we sample $\textbf{r}$ from valid rays $\hat{\mathcal{R}}$ in the updated object masks $\{\hat{M}_{n}^i\}_{n=1}^N$ after in-painting.
For the depth values, we sample $\textbf{r}$ from valid rays ${\mathcal{R}}$ in the original incomplete object masks $\{M_{n}^i\}_{n=1}^N$, because we empirically find neither the in-painted nor the predicted depth values are reliable.
The rendering-based loss can be summarized as 
\begin{equation}
\begin{split}
    \mathcal{L}_{r} = & \lambda_{\mathcal{C}} \mathop{\mathbb{E}_{\textbf{r}\in\hat{\mathcal{R}}}}(\lVert \hat{\mathcal{C}}(\textbf{r})-\mathcal{C}(\textbf{r})\lVert_1) \\
    & + \lambda_{\mathcal{N}} \mathop{\mathbb{E}_{\textbf{r}\in\hat{\mathcal{R}}}}(\lVert 1-\hat{\mathcal{N}}(\textbf{r})^T\mathcal{N}(\textbf{r})\lVert_1) \\
    & + \lambda_{\mathcal{D}} \mathop{\mathbb{E}_{\textbf{r}\in\mathcal{R}}}(\lVert \hat{\mathcal{D}}(\textbf{r})-\mathcal{D}(\textbf{r})\lVert_1),
\end{split}
\end{equation}
where $\hat{\mathcal{C}}(\textbf{r})$, $\hat{\mathcal{N}}(\textbf{r})$ and $\hat{\mathcal{D}}(\textbf{r})$ denote the rendered color, normal and depth values of ray $\textbf{r}$, respectively.

\textbf{Eikonal Loss.} For all sampled points along the ray, we add an Eikonal term \cite{eikonal} following the common practice to regularize SDF values in the 3D space

\begin{equation}
    \mathcal{L}_{eik} = \lambda_e \mathop{\mathbb{E}_{{x}\in\mathcal{X}}}(\lVert 1 -\nabla_x s(x)\lVert_1),
\end{equation}
where $\mathcal{X}$ represents the set of sampled points.

\textbf{Silhouette Loss.} Inspired by methods like GET3D \cite{get3d}, we incorporate a binary cross entropy loss for the summed weights along the ray to supervise the 3D shape from the 2D silhouette projection, which is formulated as
\begin{equation}
    \mathcal{L}_{si} = \lambda_{si} \mathcal{L}_{CE}(w(\textbf{r}), \hat{M}(\textbf{r})),
\end{equation}
where $w(\textbf{r})$ denotes the summation of weights along $\textbf{r}$ and $\hat{M}(\textbf{r})$ denotes the binary value in the updated object mask.

\begin{table*}[hbp]
    \centering
    
    \begingroup
    \begin{tabular}{ c |c c c c | c c c c}
        \specialrule{1.5pt}{1pt}{1pt}

         \multirow{2}{*}{Method}  &  \multicolumn{4}{c|}{ScanNet GT} &  \multicolumn{4}{c}{Scan2CAD GT}  \\
         \cline{2-9}
         & F-score $\uparrow$ & Acc. $\downarrow$ & Comp. $\downarrow$  &  Chamfer Dist. $\downarrow$ &  F-score $\uparrow$ & Acc. $\downarrow$ & Comp. $\downarrow$ &  Chamfer Dist. $\downarrow$  \\
        \toprule
        MonoSDF$^*$ &  0.627 & 8.60 & 11.04 & 9.82  & 0.217  & 8.18 & 14.25 & 11.22 \\
        FroDO$^*$ & 0.357 & 11.00 & 11.44  & 11.22 &  0.387 & 8.92 & 11.20 & 10.05 \\
        Scan2CAD & 0.219 & 8.05 & 20.61 & 14.33  & 0.328 & 9.05 & 19.90 & 14.45 \\
        vMap$^*$ & 0.636 & 17.47 & \textbf{3.33} & 10.40 & 0.471 & 21.17 & \textbf{5.28} & 13.23  \\
        O$^2$-Recon \textbf{(Ours)} & \textbf{0.715} & \textbf{4.32} & 4.57 & \textbf{4.45} & \textbf{0.568} & \textbf{5.96} & 6.34 & \textbf{6.15}  \\
        \specialrule{1.5pt}{1pt}{1pt}
    \end{tabular}
    \endgroup

    \caption{Evaluation of object reconstruction on the ScanNet and Scan2CAD datasets.}
    
    \label{tab:accuracy}
\end{table*}

\begin{figure*}[hbp]
  \centering 
  \includegraphics[width=\linewidth]{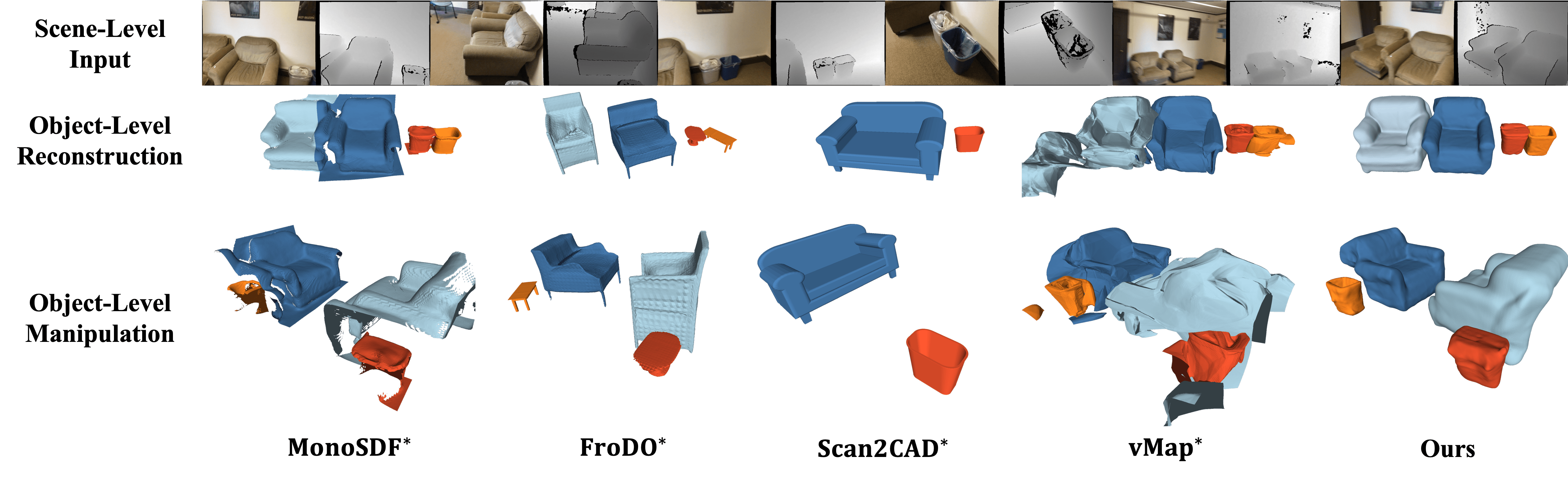}
  \vspace{-5mm}
  \caption{%
    Comparison of object-level manipulation results.
  }
  \label{fig:manipulation}
\end{figure*}

\textbf{Semantic Consistency Loss.} To improve the supervision of totally unseen areas, we apply a semantic consistency loss to the rendered color and normal images from novel views. 
Using the pre-trained CLIP \cite{clip} as encoder, we align the rendered images from novel views to the semantic space represented by the categorical text prompt $\mathcal{T}$ for in-painting and the color image $\mathcal{C}_{r}$ and normal image $\mathcal{N}_{r}$ from the reference view.
The reference view is selected by the CLIP similarity of color images and the text prompt.
Our proposed semantic consistency loss can be formulated as 
\begin{equation}
\begin{split}
    \mathcal{L}_{se} & = \lambda_{se} (\lVert 2 - \phi(\mathcal{C}_{n})^T\phi(\mathcal{T}) - \phi(\mathcal{N}_{n})^T\phi(\mathcal{T}) \lVert_1 \\
    & + \lVert 1 - \phi(\mathcal{C}_{n})^T\phi(\mathcal{C}_{r})\lVert
    + \lVert 1 - \phi(\mathcal{N}_{n})^T\phi(\mathcal{N}_{r})\lVert),
\end{split}
\end{equation}
where $\phi(\cdot)$ denotes the CLIP encoder, $\mathcal{C}_n$ and $\mathcal{N}_n$ denote the color and normal image rendered from the novel views.

Since the semantic features need to be calculated from the whole rendered images, we render $\mathcal{C}_{n}$ and $\mathcal{N}_{n}$ at a low resolution for efficiency. The semantic consistency loss is applied every several iterations to the randomly generated novel views. 
We refer readers to supplementary material for more details about the novel view generation.

\subsection{Implementation Details} 

We train our model on an NVIDIA GeForce RTX 3090 GPU for total 50k iterations using Adam optimizer, including 20k iterations for the coarse block alone and 30k iterations for both the coarse and fine blocks.
The semantic consistency loss is turned on after 10k iterations and applied every 5 iterations.
We set the initial learning rate to 2e-4 and decrease it by 0.5$\times$ every 20k iterations. 
The overall training process for one object occupies around 3GB GPU memory and takes about 4 hours.
The loss weights are set to $\lambda_{\mathcal{C}} = \lambda_{\mathcal{N}} = \lambda_{\mathcal{D}} = \lambda_{se} = 1.0$, $\lambda_e = 0.1$, and $\lambda_{si} = 5.0$.

\section{Experiments}

\subsection{Datasets, Baselines and Evaluation Metrics}

\noindent\textbf{Datasets.} We evaluate O$^2$-Recon on 6 scenes from ScanNet \cite{scannet}, encompassing 77 objects.  A significant portion of these objects are occluded, presenting difficulties in achieving complete reconstructions.
We follow the practice in vMap \cite{vmap} to get the segmentation results of the scene-level videos.
Since the 3D ground truth of occluded regions are missing in the ScanNet, we leverage the aligned CAD models in Scan2CAD \cite{scan2cad} to evaluate the reconstruction accuracy of unseen regions.
In our experiments, 59\% of the occluded objects require 1 user-sketched mask, 29\% require 2 user-sketched masks, and 12\% require 3 user-sketched masks.

\noindent\textbf{Baselines.} We compare our method with the following state of the arts. 
(1) The scene-level reconstruction method MonoSDF \cite{monosdf}. We leverage the ground truth depth in its optimization and denote it as MonoSDF$^*$.
(2) The re-implemented shape-code-based method FroDO \cite{frodo}, denoted by FroDO$^*$. 
(3) The Scan2CAD method \cite{scan2cad} based on CAD model retrieval.
(4) The general object-level reconstruction method vMap \cite{vmap}. We optimize it with more iterations for fair comparison and denote it as vMap$^*$.
We refer readers to SM for more details about the baseline methods.

\noindent\textbf{Metrics.} We follow the previous work \cite{vmap} to evaluate the reconstruction accuracy with the F-score within 5cm and the Chamfer distance (cm).
We also report the accuracy and completion terms 
for detailed analysis.

\begin{figure*}[!htbp]
  \centering 
  \includegraphics[width=0.98\linewidth]{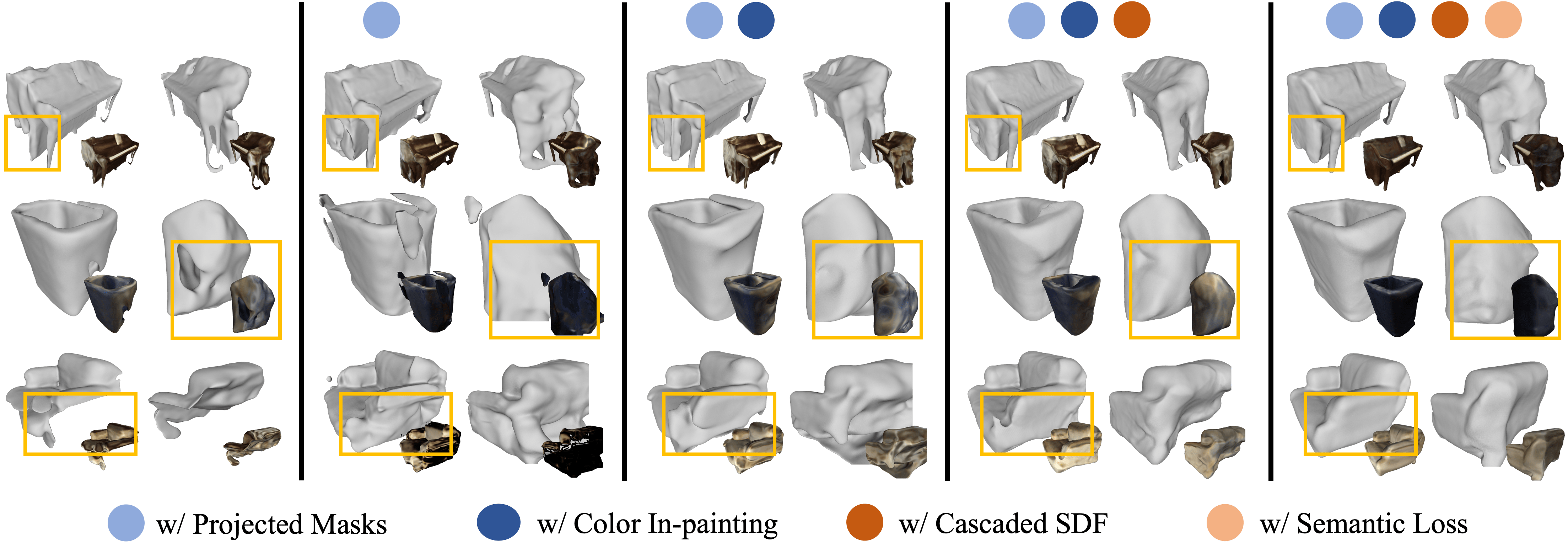}
  \caption{%
    The ablation studies demonstrate how our carefully designed components progressively fill in the occluded regions.
  }
  \label{fig:ablation}
\end{figure*}

\subsection{Comparisons}

\noindent\textbf{Qualitative Evaluation.} Figure \ref{fig:main-exp} compares the reconstruction results of different methods on objects that suffer from various occlusion conditions. 
Notably, the scene-level method MonoSDF$^*$ cannot reconstruct complete surfaces for occluded regions, and sometimes fails on certain cases, e.g., the last row.
As for the FroDO$^*$ method based on shape code, although complete meshes can be derived from the latent space, it cannot match the actual surface very well, and cannot reconstruct 3D objects of arbitrary categories, e.g., the piano and the trash bin.
The Scan2CAD method can retrieve proper CAD models from the database, but the optimized scale and pose parameters are often unsatisfactory. 
The NeRF-based method vMap$^*$ generates accurate surfaces for visible areas of arbitrary objects, but produces holes or degenerated artifacts in the unseen areas.
In contrast, our proposed system O$^2$-Recon simultaneously reconstructs accurate surfaces for visible regions and plausible surfaces for invisible regions. 
We ensure the accuracy of visible surfaces by rendering-based SDF optimization and the plausibility of unseen surfaces by the 2D in-painting.

\noindent\textbf{Quantitative Evaluation}.
We use the ground truth in ScanNet and Scan2CAD datasets to evaluate the accuracy of the object-level reconstructions.
The ScanNet ground truth contains accurate surfaces fused from depth inputs, which can be utilized to measure the accuracy of visible regions.
As for the occluded areas, we measure the accuracy and plausibility using ground truth in the Scan2CAD dataset, which contains annotated CAD models for objects in the scenes that are complete and roughly match the actual objects.

As shown in Table \ref{tab:accuracy}, our method outperforms all baseline methods in terms of the overall F-score and Chamfer distance. 
Compared to the baseline methods, our method reduces the Chamfer distance by around 50\% and improves the F-score by more than 10\%.
We also notice that vMap performs better in the completion term but receives the largest error in the accuracy term, since it reconstructs a lot of surfaces in the empty space, as shown in Figure \ref{fig:main-exp}.
These quantitative results are consistent with our qualitative analysis, and demonstrate the superiority of our proposed method.

\begin{table}[t]
    \centering

    \begingroup
    \setlength{\tabcolsep}{3.5pt}
    \begin{tabular}{ l | c c c c c}
        \specialrule{1.5pt}{1pt}{1pt}
          & V1 & V2 & V3 & V4 & Full     \\
         \toprule
          w/ Projected Masks &  & \Checkmark & \Checkmark & \Checkmark & \Checkmark   \\
          w/ Color In-painting &  &  & \Checkmark &  \Checkmark & \Checkmark  \\
          w/ Cascaded SDF &  &  &  & \Checkmark & \Checkmark    \\
          w/ Semantic Loss &  &  &  &  & \Checkmark   \\
          \toprule
          ScanNet F-score $\uparrow$ & 0.718 & 0.697 & \textbf{0.720} & 0.713 & 0.715    \\
          ScanNet C.D. $\downarrow$ & 4.70 & 5.17 & 4.65 & 4.53 & \textbf{4.45}     \\
          \toprule
          Scan2CAD F-score $\uparrow$ & 0.502 & 0.511 & 0.553 & 0.560 & \textbf{0.568}     \\  
          Scan2CAD C.D. $\downarrow$ & 9.73 & 8.72 & 6.80 & 6.51 & \textbf{6.15}    \\

      \specialrule{1.5pt}{1pt}{1pt}
                
    \end{tabular}
    \endgroup
    
    \caption{Quantitative results for the ablation studies. 
    }
    
    \label{tab:ablation}
\end{table}

\noindent\textbf{Object-Level Manipulation.}
Based on the independent reconstructed objects, we can achieve object-level manipulation with few artifacts due to the high accuracy and completeness of O$^2$-Recon.
As shown in Figure \ref{fig:manipulation}, 3D reconstructions generated by O$^2$-Recon maintain a good visualization effect after large-scale manipulation. 
While the 3D manipulation results based on other methods contain artifacts like missing or floating parts and inaccurate geometry.

\subsection{Ablation Study}

We evaluate the impact of different components on achieving complete 3D reconstruction by comparing our full method to four other variants as shown in Table \ref{tab:ablation}.

\noindent\textbf{Color In-painting.} The 2D color in-painting is a crucial component in our method. While we can already reconstruct some occluded areas using the projected masks without color in-painting, these masks are not precise enough. This lack of precision leads to surfaces that can look distorted or noisy, as seen in the second column of Figure \ref{fig:ablation}. By adopting color in-painting followed by mask refinement, we produce more accurate shapes for obscured areas. This is evident in the third column of Figure \ref{fig:ablation}. The quantitative results reported in Table \ref{tab:ablation} also confirms the effectiveness of the color in-painting step for improving both the F-score and Chamfer distance measures.

\noindent\textbf{Cascaded SDF Architecture and Semantic Loss.}
To enhance the reconstruction of completely hidden areas, we introduce a cascaded SDF architecture coupled with a semantic consistency loss. Both strategies, as illustrated in the last two columns of Figure \ref{fig:ablation} and Table \ref{tab:ablation}, contribute to a smoother and more precise reconstructed surface in these unseen regions. In particular, the supervision provided by the semantic consistency loss significantly boosts the color consistency of the resulting surface.

\section{Conclusion }

In this paper, we introduce O$^2$-Recon for reconstructing complete 3D geometry of occluded objects in a scene using a pre-trained 2D diffusion model. 
We utilize the diffusion model to in-paint the occluded parts in multi-view 2D images, and then reconstruct 3D objects using neural implicit surface from the in-painted images. To prevent inconsistency in mask generation, we adopt a human-in-the-loop strategy that can effectively guide the 2D in-painting process with only a few human interaction.
During the optimization process of neural implicit surfaces, we design a cascaded SDF architecture to guarantee smoothness, and also leverage the pre-trained CLIP model to supervise novel views with semantic consistency loss.
Our experiments on the ScanNet scenes show that O$^2$-Recon is able to reconstruct accurate and complete 3D surfaces for occluded objects from any category. 
The reconstructed 3D objects can be utilized in further manipulation like large rotations and translations.

\clearpage

\setcounter{secnumdepth}{0}
\section{Acknowledgments}
This work was partially supported by the Fundamental Research Funds for the Central Universities (2023XKRC045), the Natural Science Foundation of China (Project Number U2336214) and the Key Laboratory of Pervasive Computing, Ministry of Education, China.

\bibliography{aaai24}

\clearpage
\setcounter{secnumdepth}{2}

\renewcommand{\thetable}{S\arabic{table}}
\renewcommand{\thefigure}{S\arabic{figure}}
\renewcommand{\thesection}{S\arabic{section}}
\renewcommand{\theequation}{S\arabic{equation}}

%

\setcounter{page}{1}
\setcounter{table}{0}
\setcounter{section}{0}
\setcounter{equation}{0}
\setcounter{figure}{0}
\maketitlesupplementary

\section{The Tool for User Sketching}

\begin{figure}[h]
  \centering 
  \includegraphics[width=\linewidth]{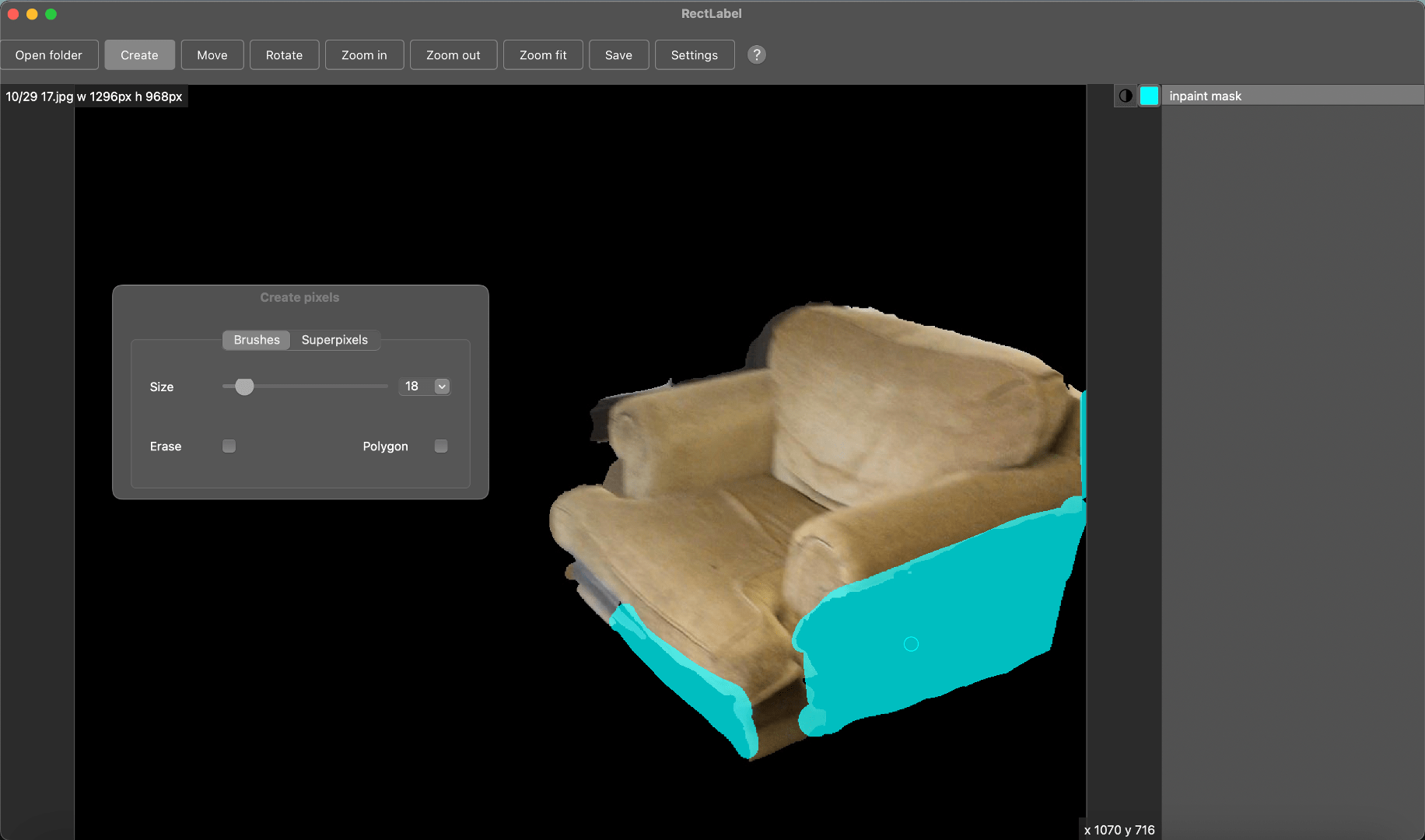}
  \caption{%
     A screenshot of the user sketching GUI.
  }
  \label{fig:rectlabel}
\end{figure}

We use the RectLabel software \cite{rectlabel} as a tool to facilitate the user sketching step. The interface, as depicted in Figure \ref{fig:rectlabel}, enables users to conveniently navigate through all the RGB images associated with an object. By utilizing the brush tool provided by RectLabel, users can perform segmentation annotation, sketching the areas that require in-painting based on their expertise and experience.

\section{Masks Sketched by Different Users} 

To study the influence of masks sketched by different users, we ask 5 people to select frames and draw in-painting masks for all the 77 objects. 
We evaluate the reconstruction results on Scan2CAD annotations and report the statistics in Table \ref{tab:human}. The results show that our system consistently reconstructs 3D meshes with good quality given different sketched masks of different selected frames.

\begin{table}[t]
    \centering
    
    \begingroup
    \setlength{\tabcolsep}{5.5pt}
    \begin{tabular}{ l | c c c c}
        \specialrule{1.5pt}{1pt}{1pt}
          & Mean & Std & Max &  Min    \\
         \toprule
          F-score [\textless 5cm \%] $\uparrow$ & 0.573 & 0.016 & 0.588 &   0.550   \\
          \toprule
          Acc. Dist. [cm] $\downarrow$ & 5.79 & 0.36 & 6.30 & 5.37     \\
          Comp. Dist. [cm] $\downarrow$ & 6.37 & 0.32 & 0.68 & 0.60     \\
          Chamfer Dist. [cm] $\downarrow$ & 6.08 & 0.11 & 6.20 & 5.91    \\

      \specialrule{1.5pt}{1pt}{1pt}
                
    \end{tabular}
    \endgroup

    \caption{Reconstruction accuracy statistics of O$^2$-Recon with different user engagement. }
    
    \label{tab:human}
\end{table}

\begin{figure}[t]
  \centering 
  \includegraphics[width=\linewidth]{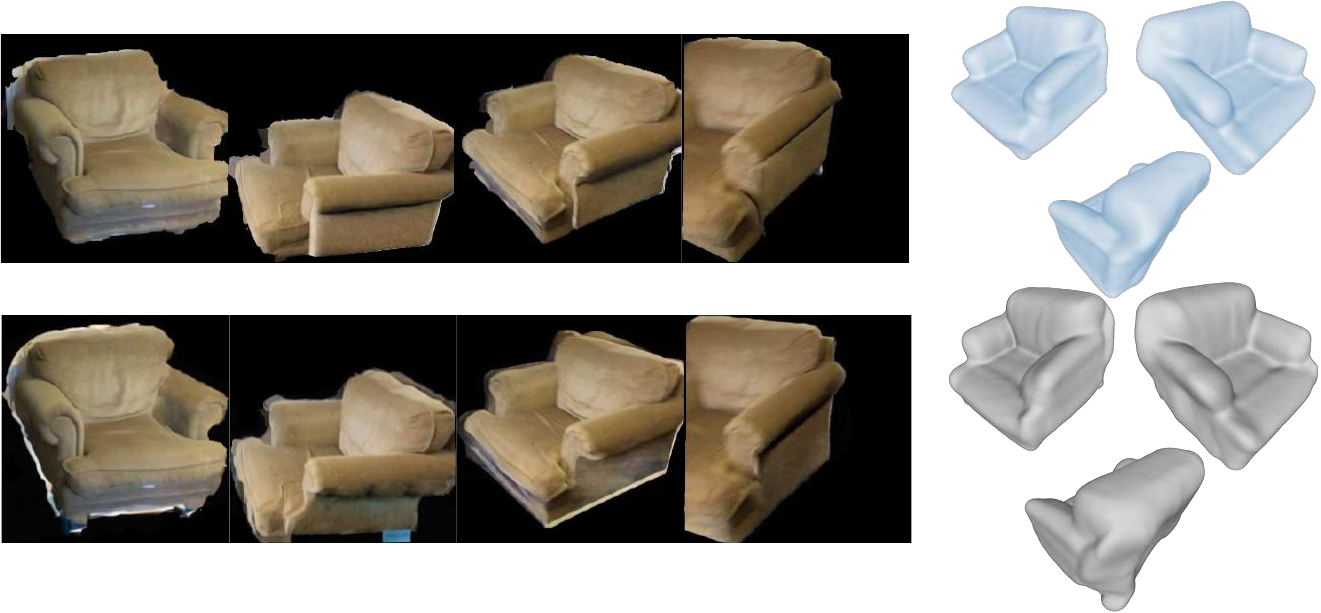}
  \caption{%
     Comparison of surface reconstruction from all user-sketched in-painting masks (top) and inpainting masks generated from our pipeline (bottom).
  }
  \label{fig:ablation-2}
\end{figure}


\section{Comparison with All User-Sketched Masks} 
To validate the effectiveness of our proposed human-engaged mask generation process, we conducted a comparison with high-quality all human-sketched in-painting masks. 
The results of this comparison are depicted in Figure \ref{fig:ablation-2}.
In the figure, we can observe that both the quality of the in-painted images and the reconstructed 3D mesh are comparable to those obtained using all-user-sketched masks.
This indicates that our semi-automatic in-painting process yields satisfactory results with significantly less cost of labor and time.

\section{Depth In-painting and Mask Projection}

In the mask projection step, our goal is to propagate the in-painting masks from the selected views to all other views.

\begin{figure}[h]
  \centering 
  \includegraphics[width=\linewidth]{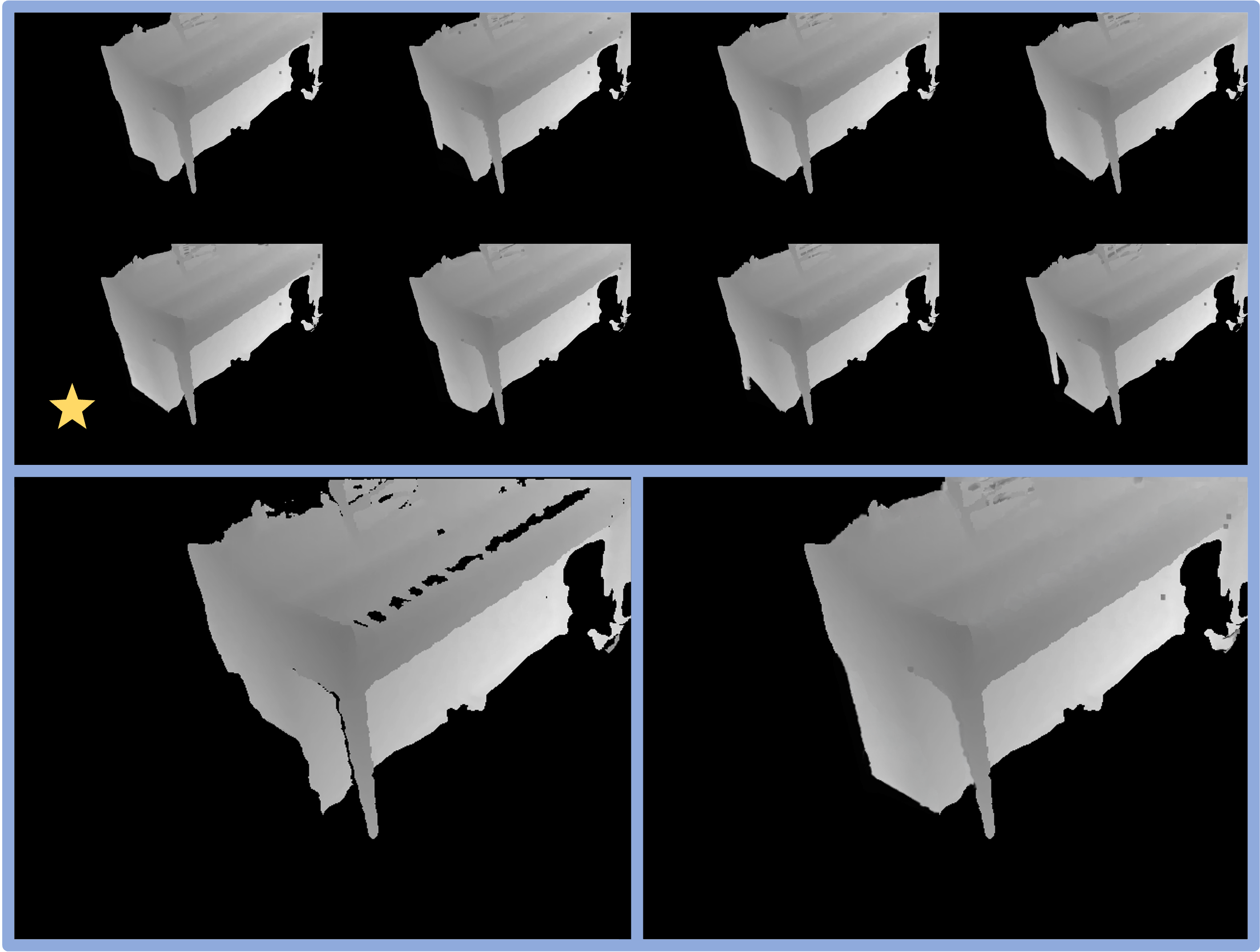}
  \caption{%
     Middle and final results of the depth in-painting step. The upper part shows 8 in-painting results, and the lower part shows the incomplete input depth and the produced complete depth result.
  }
  \label{fig:depth-inpainting}
\end{figure}

To propagate the mask information through re-projection, we utilize the diffusion model to in-paint depth values within the sketched area.
We leverage the Stable Diffusion in-painting model \cite{stable-diffusion} as the diffusion model in our implementation.
The depth maps are firstly normalized from $[0, D_{max}]$ to $[0,255]$, where $D_{max}$ denotes the maximum depth value.
Treating the normalized depth maps as grayscale images, we input them into the diffusion-based in-painting model.
After undergoing the in-painting process, the in-painted gray images are then scaled back to $[0, D_{max}]$ and form the in-painted depth maps.
To resist the randomness of the diffusion model and ensure the quality of in-painted depth values, we in-paint each selected view 8 times and select the output with the largest valid area.
Since we have only 1-3 selected views per instance, the time cost consumed by the repeated in-painting process is acceptable.
As shown in Figure \ref{fig:depth-inpainting}, our repeat-and-select process is effective and can produce good in-painted depth maps.

Using the depth information predicted by the diffusion model, we can perform a back-projection of the mask pixels from the 2D images to the corresponding 3D space. 
This process generates a point cloud that encompasses the mask pixels from all the selected views. 
Subsequently, these 3D points are projected back onto the 2D images of all other views.
However, it is worth noting that the point cloud projection often leads to sparsity in the resulting 2D images.
Moreover, the presence of inaccurate depth predictions can lead to incorrect projections, resulting in isolated points that fall outside the subject area.
To obtain good in-painting masks, we apply morphological processing operations after the mask projection step.
These operations facilitate refinement and improvement of the masks by filling in sparse regions and eliminating isolated points that fall outside the subject area.

Specifically, after the projection of 3D points, we firstly discard the pixels that have less than 5 positive neighbors in their 3x3 neighborhood.
Such operation filters the noisy isolated points outside the subject area.
To connect the sparse components within the subject area, we then apply a dilation operation to the filtered mask image.
This dilation process helps bridge gaps and fills in missing regions, resulting in solid and connected regions. 
These regions can then be interpreted as reliable masks for further processing.
To avoid modifications to the visible regions, we then subtract the original object mask from the dilated areas. 
To address potential pixel-level error in the instance masks, we perform an additional dilation operation after the subtraction step, and leverage the results as in-painting masks for the diffusion model.

With the generated in-painting masks, the pre-trained diffusion model is able to plausibly in-paint the occluded regions of objects.
However, due to variations in neighborhood information across different pixels, certain regions within the in-painting masks may be filled with \textit{black pixels} by the diffusion model.
It is important to note that these \textit{black pixels} cannot be considered as valid areas and should be excluded from the instance masks of the in-painted images.
Thus we update the instance masks after in-painting by excluding all-zero pixels.
This post-inpainting mask update not only eliminates the \textit{black pixels} but also helps filter out any remaining noisy regions that may have resulted from the mask projection step. 

\begin{figure}[h]
  \centering 
  \includegraphics[width=\linewidth]{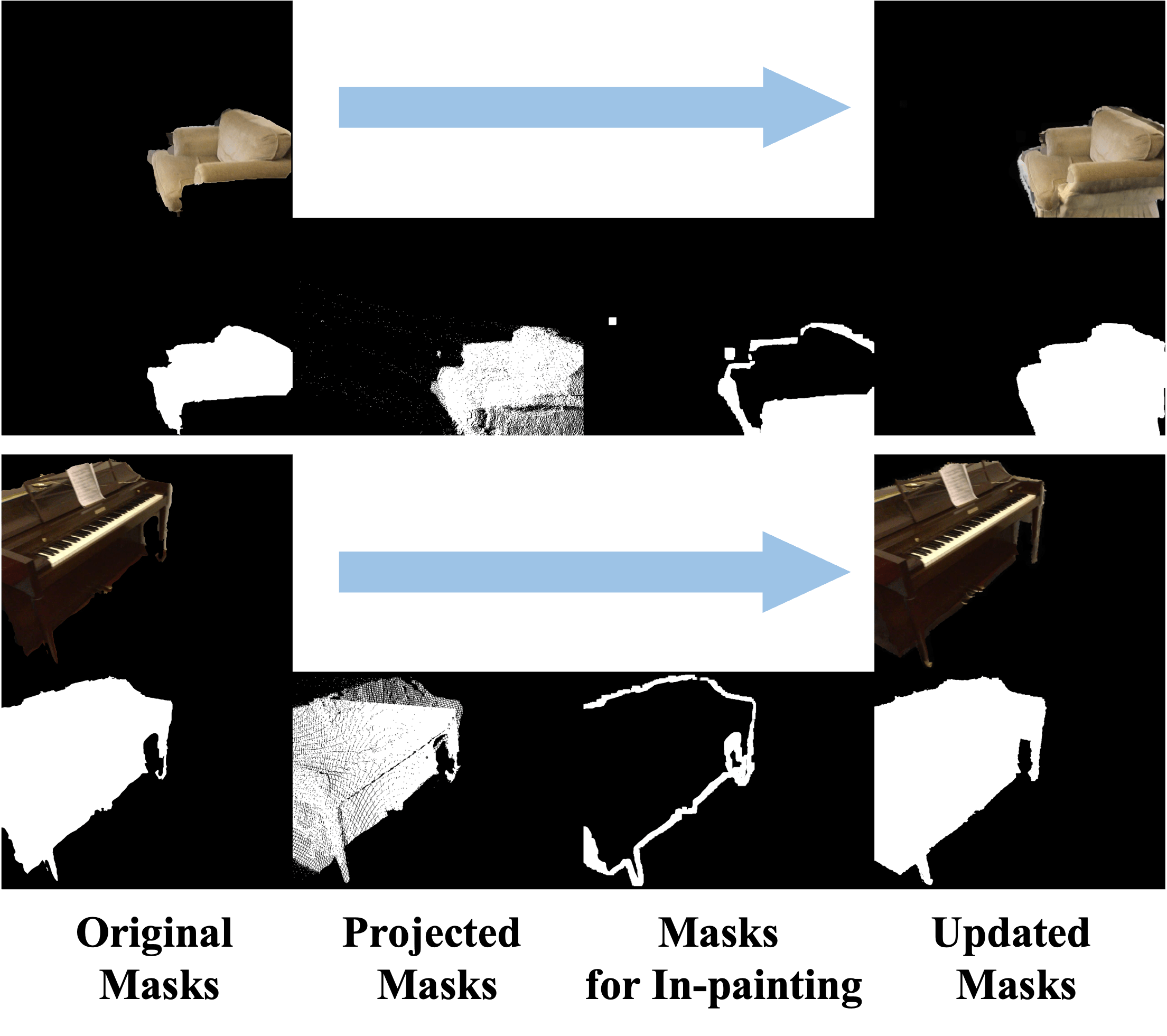}
  \caption{%
     The middle results of in-painting masks for views that are not selected.
  }
  \label{fig:mask-dilation}
\end{figure}

Figure \ref{fig:mask-dilation} illustrates the evolving process of object masks throughout the entire in-painting procedure. 
The final instance masks exhibit significant improvements compared to the original occluded masks, offering more comprehensive guidance for the geometry optimization process.
By applying various operations such as dilation, subtraction, and post-inpainting mask updating, we refine the object masks to ensure better alignment with the completed regions.
These refined instance masks play a crucial role in providing accurate and detailed supervision during the optimization of the geometry, ultimately leading to more visually pleasing and faithful reconstructions.

\section{Novel View Generation}

To effectively supervise the implicit representation from novel views, we first randomly select novel camera poses based on the captured views, and then render the low-resolution outputs from that viewpoint for efficiency.
In this section, we describe our novel view selection strategy and the techniques for fast rendering during the training process.

For iterations that requires the semantic consistency supervision, we randomly choose a reference camera pose from the captured viewpoints.
However, since most reference poses are not oriented towards the center of the object, we perform a correction step to ensure that a significant portion of the object falls within the center region of the novel view.
To correct the view direction of the reference pose, we set it as a vector pointing towards the center of the object from the reference pose's position. 
The center point coordinates can be obtained from the point cloud of the object, which is generated by fusing the masked depth frames.
Next, we generate a novel camera pose near the corrected reference pose. 
Specifically, we parameterize the camera pose using three parameters: yaw angle, pitch angle, and radius with respect to the center point of the object. 
The angle parameters are sampled from normal distributions with predefined standard deviations, where the mean values are set to the corresponding parameters of the corrected reference pose. 
The radius of the novel pose is sampled from a uniform distribution, with the maximum and minimum values determined as follows:
\begin{equation}
    r_{min} = \min(\frac{D_{bbox}}{2}, \{d_i\}_{\min}),
\end{equation}
\begin{equation}
    r_{max} = \max(\frac{D_{bbox}}{2}, \{d_i\}_{\max}) \times 0.9,
\end{equation}
where $D_{bbox}$ denotes the diagonal length of the object point cloud's 3D bounding box and $d_i$ denotes the distance between the object center and the position of viewpoint $i$.
With the sampled parameters, we calculate the camera pose of novel viewpoint, and leverage it to render the RGB images and surface normal maps.

\begin{figure}[h]
  \centering 
  \includegraphics[width=\linewidth]{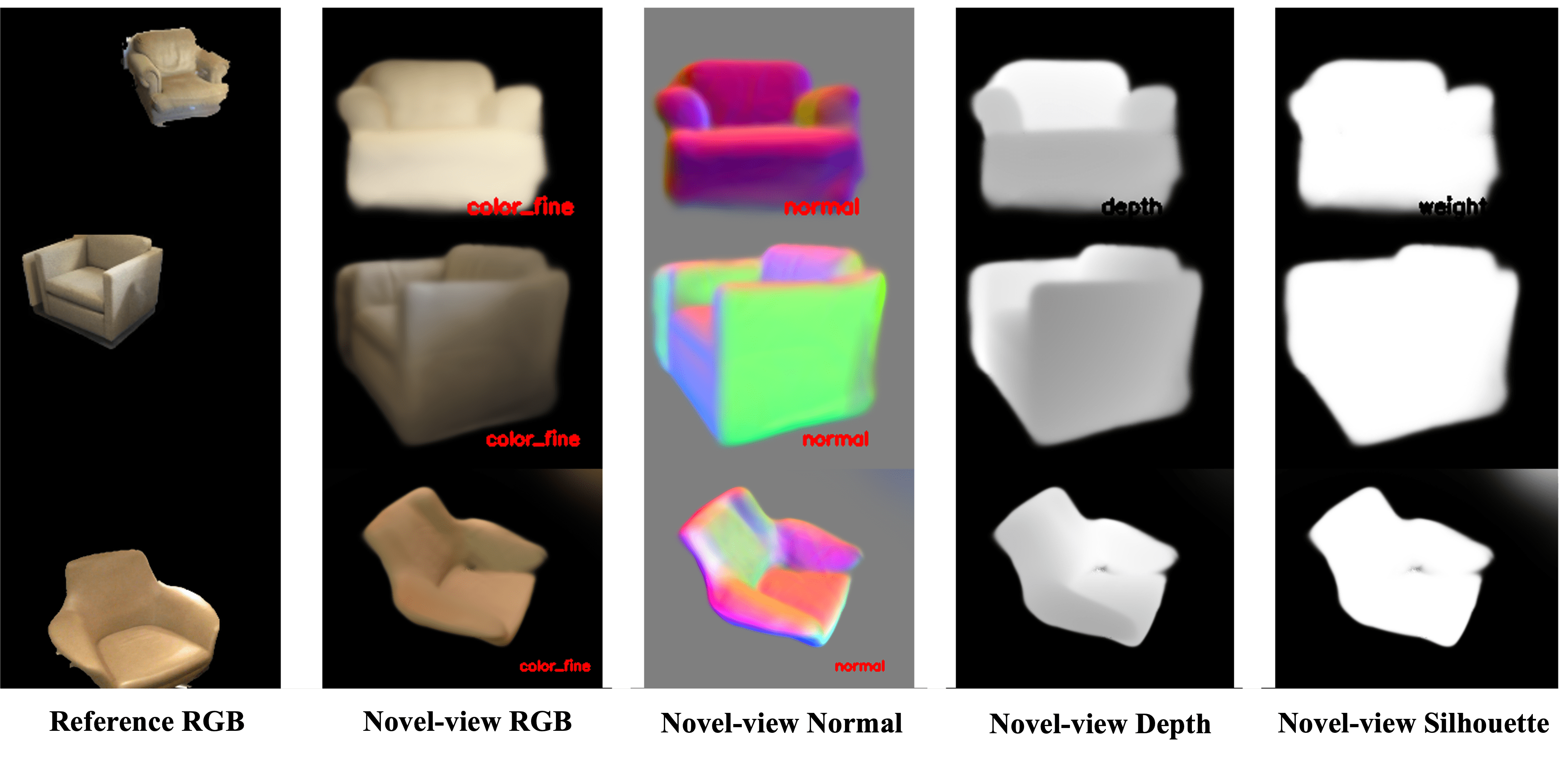}
  \caption{%
     The reference image and renderings from novel views.
  }
  \label{fig:novel-view}
\end{figure}

To ensure efficient training with semantic consistency supervision, we employ two techniques to accelerate the rendering process: region of interest localization and rendering resolution reduction. 
Both techniques effectively reduce the number of pixels to be rendered, improving rendering efficiency and training speed. 
In particular, we firstly project corner points of the object point cloud's 3D bounding box onto the novel view to approximate its location in the 2D image.
From these projected points, we extract an axis-aligned 2D bounding box that represents the region of interest.
By focusing on this smaller region, we only need to render a cropped patch instead of the entire image, significantly reducing computation.
Moreover, we also down-sample the rendering resolution to further reduce the rendering burden and improve the encoding speed of CLIP.
In Figure \ref{fig:novel-view}, we provide a visualization of the RGB image at the reference pose and the renderings obtained from the randomly sampled novel view using the aforementioned techniques.
This demonstrates how these techniques contribute to efficient rendering and training in our approach.

\section{Implementation Details} 

Our proposed model is experimented on an NVIDIA GeForce RTX 3090 GPU. 
For the color branch, we adopt a 2-layer MLP with 64 hidden units, which is much smaller than the scene-level configuration in NeuRIS \cite{neuris}.
As for the cascaded SDF branch, we set the frequency of positional encodings for the coarse and fine blocks as $0.5\times$ and $1.0\times$ of that in NeuS \cite{neus}, respectively. 
We use a four-layer MLP for the coarse block and another two-layer MLP for deeper layers of the fine block, with the same 64 hidden units as the color branch.

We implement model in Pytorch and train it using Adam optimizer.
The loss weights are set to $\lambda_{\mathcal{C}} = \lambda_{\mathcal{N}} = \lambda_{\mathcal{D}} = \lambda_{se} = 1.0$, $\lambda_e = 0.1$, and $\lambda_{si} = 5.0$. 
The total training process takes 50k iterations. 
Specifically, we train the coarse block alone for 20k iterations and then together with the fine block for another 30k iterations.
The semantic consistency loss is turned on after 10k iterations and applied every 5 iterations.
And the initial learning rate is set to 2e-4 and decreases by 0.5 every 20k iterations. 
We randomly select 512 rays for each iterations and sample 64 points on each ray.
The overall training process for one object occupies around 3GB GPU memory and takes about 4 hours. 

\section{Details of the Baseline Methods}

We compare O$^2$-Recon with four baselines, each representing a type of methods that can be utilized in object-level 3D reconstruction.
The baseline methods are described as follows.
\begin{itemize}
    \item The scene-level reconstruction method MonoSDF \cite{monosdf}. MonoSDF is based on implicit surface in the form of MLP layers and utilizes geometry priors to improve the reconstruction accuracy. Instead of the depth information estimated from monocular images, we leverage the groundtruth depth to optimize the MLP for fair comparison and denote it as MonoSDF$^*$.
    \item The most representative shape code based method FroDO \cite{frodo}. Since most of the shape code based methods, like \cite{moltr,centersnap,ellip-sdf}, are not open-source, we re-implement the representative FroDO method and denote it as FroDO$^*$. We integrate the FroDO models of two categories, tables and chairs, to output the final results. 
    \item The Scan2CAD method \cite{scan2cad} based on CAD databased retrieval. We follow the instructions in the officially released codebase and evaluate this method.
    \item The general object-level reconstruction method vMap \cite{vmap}. Since vMap is proposed for real-time reconstruction and is trained with limited iterations, we increase its training iteration to the same as O$^2$-Recon for fair comparison and denote it as vMap$^*$.
\end{itemize}

\section{Comparison with the 3D Completion Method}

\begin{figure}[h]
  \centering 
  \includegraphics[width=\columnwidth]{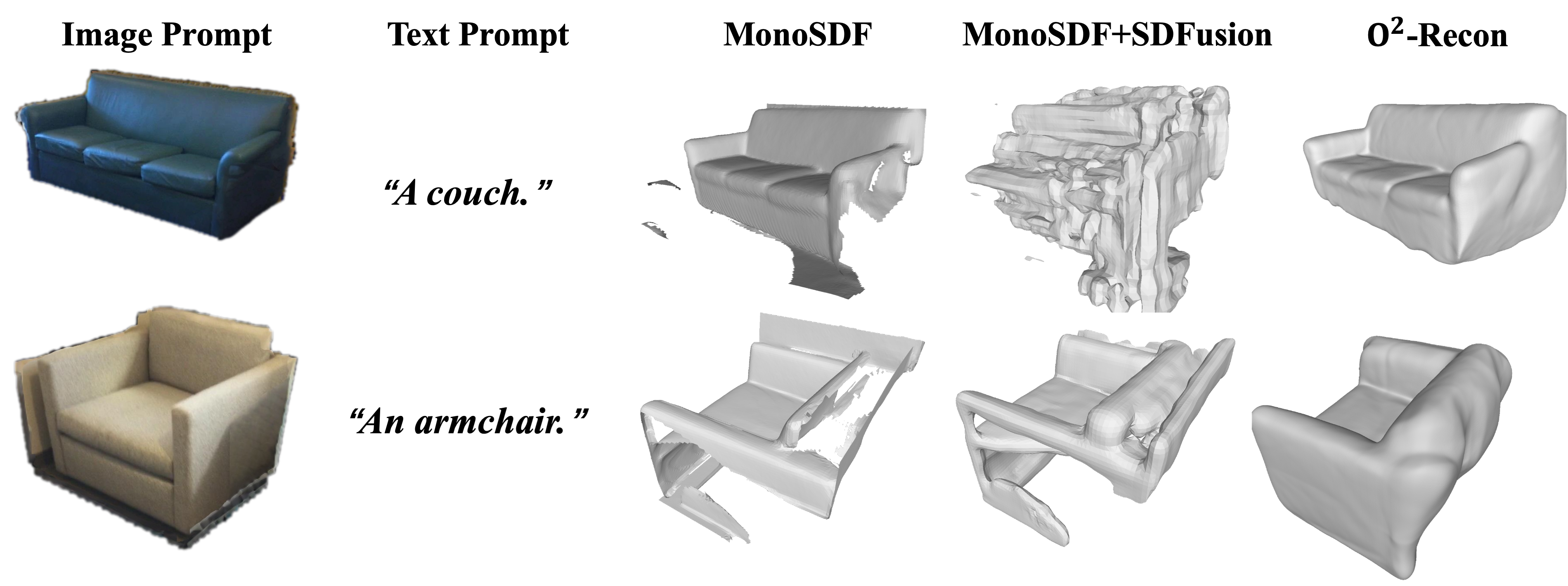}
  \caption{%
     3D completion results of SDFusion conditioned on an in-painted image and the categorical text prompt.
  }
  \label{fig:sdfusion}
\end{figure}

We utilize a recent 3D completion method, SDFusion (Cheng et al., CVPR 2023), to complete the 3D surface reconstructed by MonoSDF conditioned on image and text prompts. The 3D completion mask is set to $|SDF(x)|>1/32$ for the $64^3$ SDF volume. 
The results presented in Table \ref{tab:rebuttal-exp} and Figure \ref{fig:sdfusion} clearly show that this method cannot produce reasonable geometry for the occluded parts and even compromises MonoSDF's performance. The core issue stems from an overreliance on highly idealized training datasets such as ShapeNet, which fails to capture the complexity of real-world scenarios.

\begin{table}[h]
    \small
    \centering
    
    \begingroup
    \setlength{\tabcolsep}{3.5pt}
    \begin{tabular}{ l |c c | c c}
        \specialrule{1.5pt}{1pt}{1pt}

         \multirow{2}{*}{Method}  &  \multicolumn{2}{c|}{ScanNet GT} &  \multicolumn{2}{c}{Scan2CAD GT}  \\
         \cline{2-5}
         & F-score $\uparrow$ &  C.D. $\downarrow$ &  F-score $\uparrow$ &  C.D. $\downarrow$  \\
        \toprule
        MonoSDF & 0.627 & 9.82 & 0.217 & 12.22\\
        MonoSDF + SDFusion & 0.509 & 12.02 & 0.169 & 12.05\\
        \toprule
        Ours w/o $\mathcal{T}$ & 0.715 & 4.50 & 0.562 & 6.32\\
        \textbf{Ours} & \textbf{0.715} & \textbf{4.45} & \textbf{0.568} & \textbf{6.15}  \\
        \specialrule{1.5pt}{1pt}{1pt}
    \end{tabular}
    \endgroup

    \caption{Additional results of geometry evaluation. 
    }
    
    \label{tab:rebuttal-exp}
    
\end{table}

\section{More Visualizations of the Reconstruction Results}

\begin{figure*}[t]
\centering

     \begin{subfigure}[b]{1.0\linewidth}
         \centering
         \includegraphics[width=1.0\linewidth]{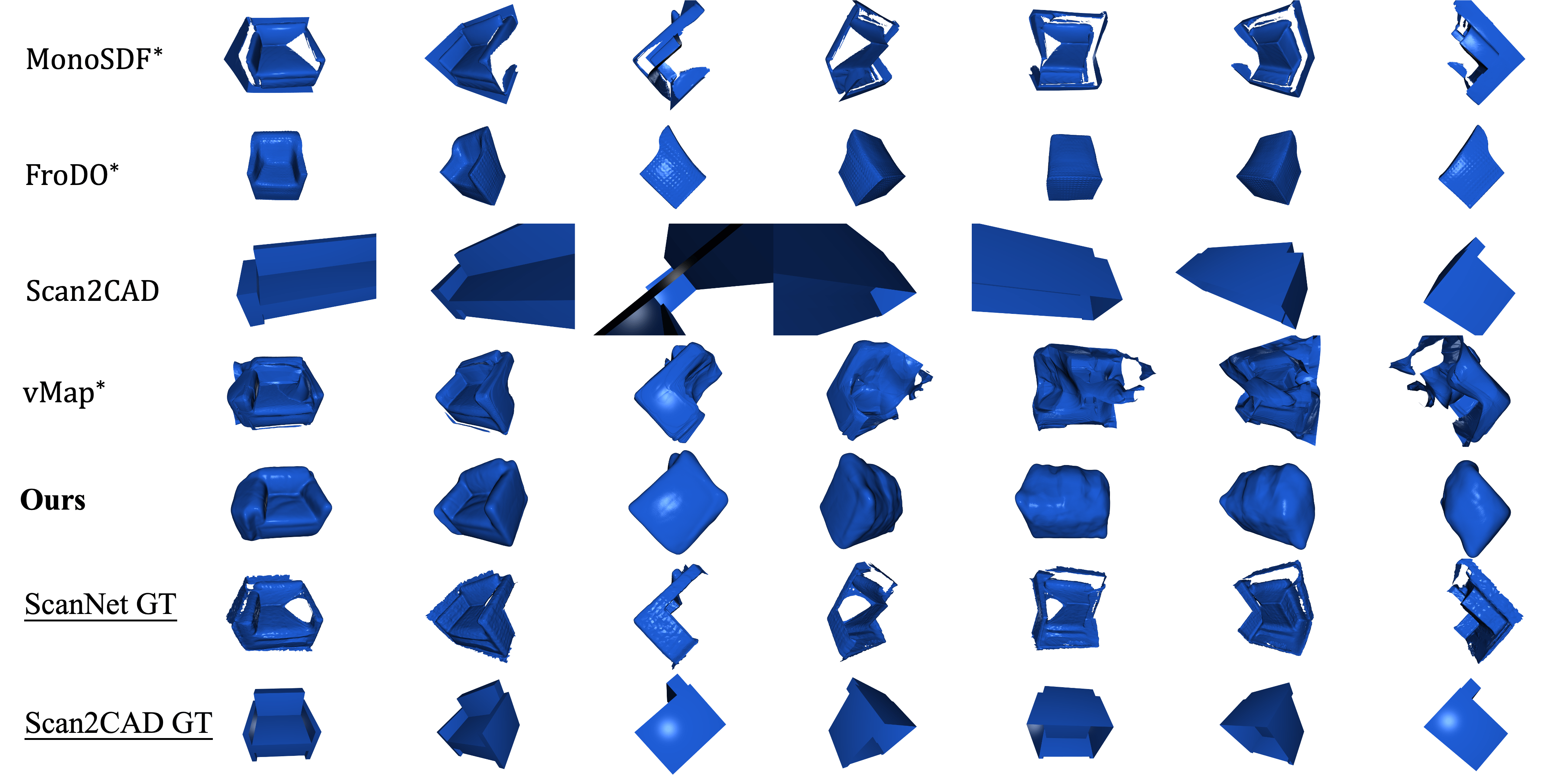}
         \label{fig:vis-obj1}
     \end{subfigure}
     \\[\smallskipamount]
     \begin{subfigure}[b]{1.0\linewidth}
         \centering
         \includegraphics[width=1.0\linewidth]{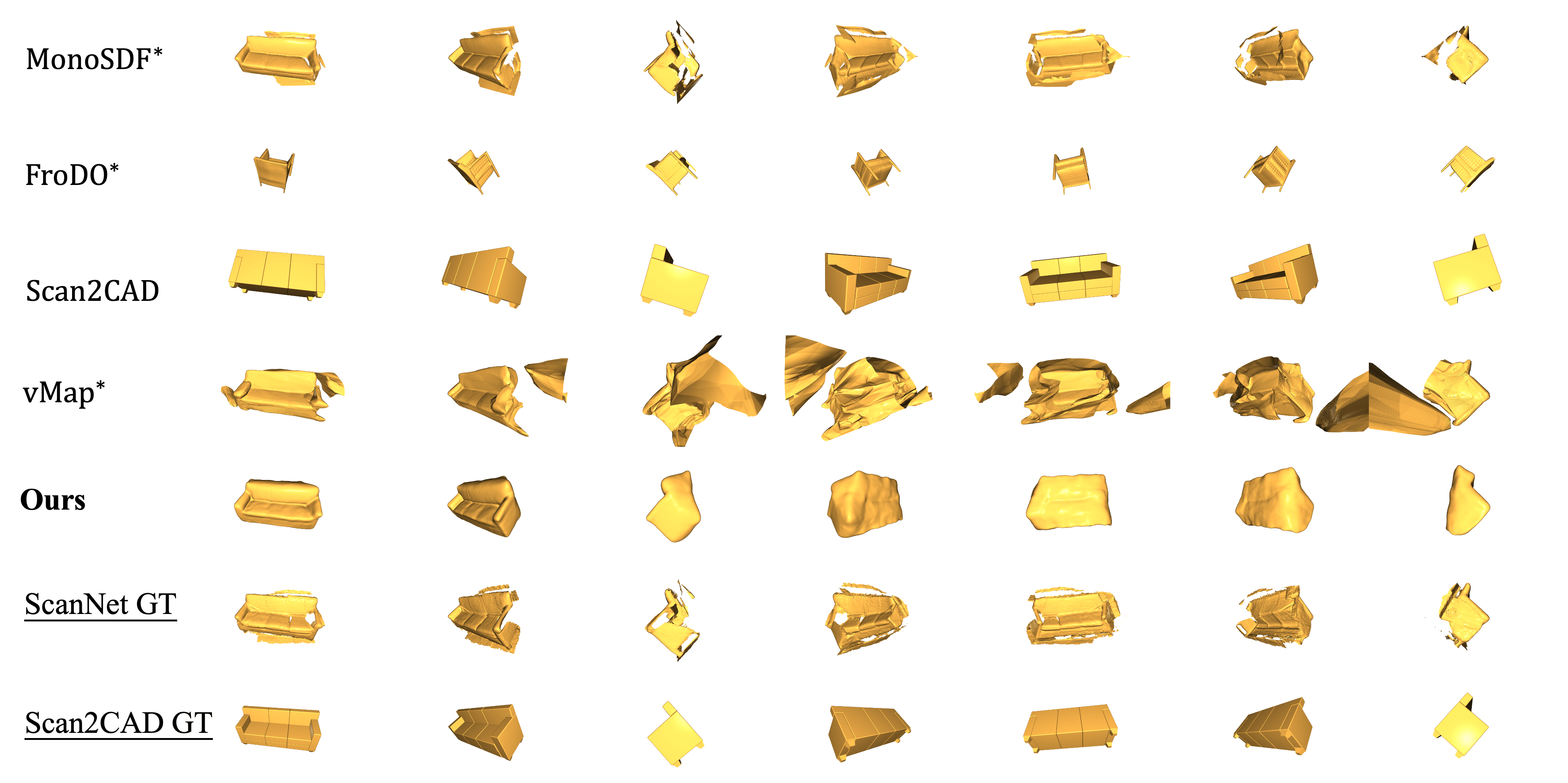}
         \label{fig:vis-obj12}
     \end{subfigure}

\caption{More visualizations of the qualitative comparisons (1-2).
}
\label{fig:visualization-1}
\end{figure*}

\begin{figure*}[t]
\centering
     \begin{subfigure}[b]{1.0\linewidth}
         \centering
         \includegraphics[width=1.0\linewidth]{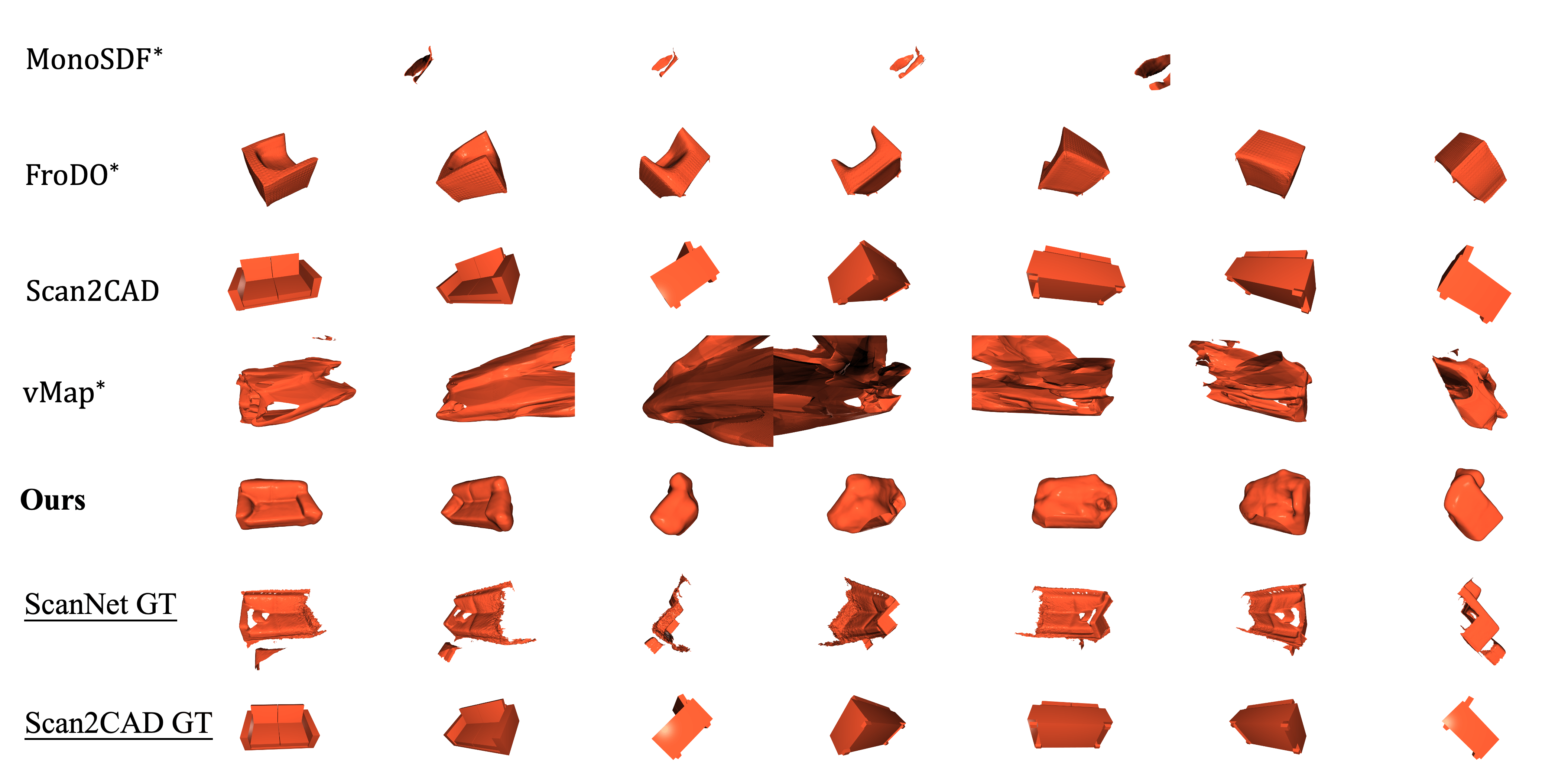}
         \label{fig:vis-obj5}
     \end{subfigure}
     \\[\smallskipamount]
     \begin{subfigure}[b]{1.0\linewidth}
         \centering
         \includegraphics[width=1.0\linewidth]{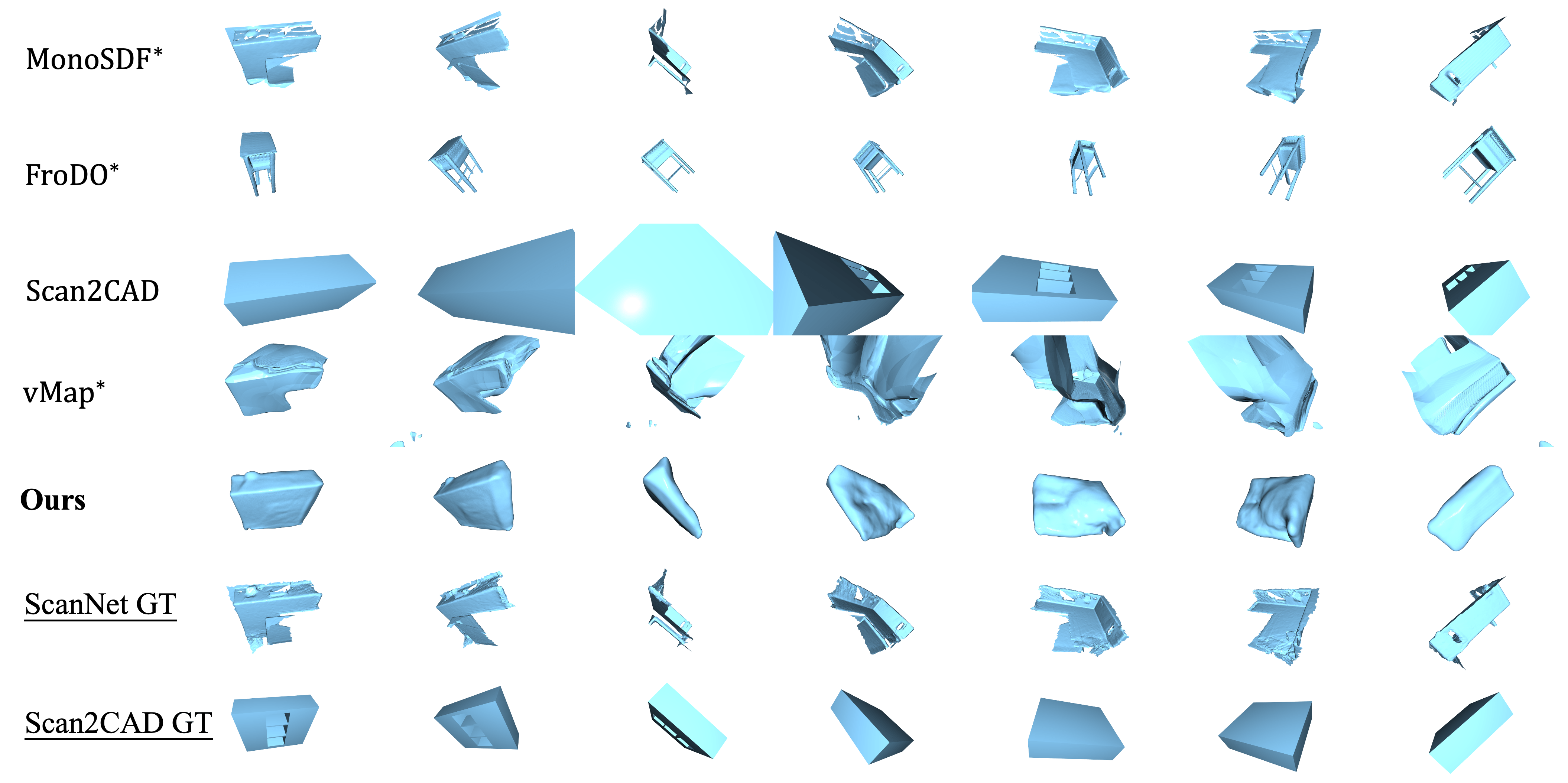}
         \label{fig:vis-obj6}
     \end{subfigure}

\caption{More visualizations of the qualitative comparisons  (3-4).
}
\label{fig:visualization-2}
\end{figure*}

\begin{figure*}[t]
\centering
     \begin{subfigure}[b]{1.0\linewidth}
         \centering
         \includegraphics[width=1.0\linewidth]{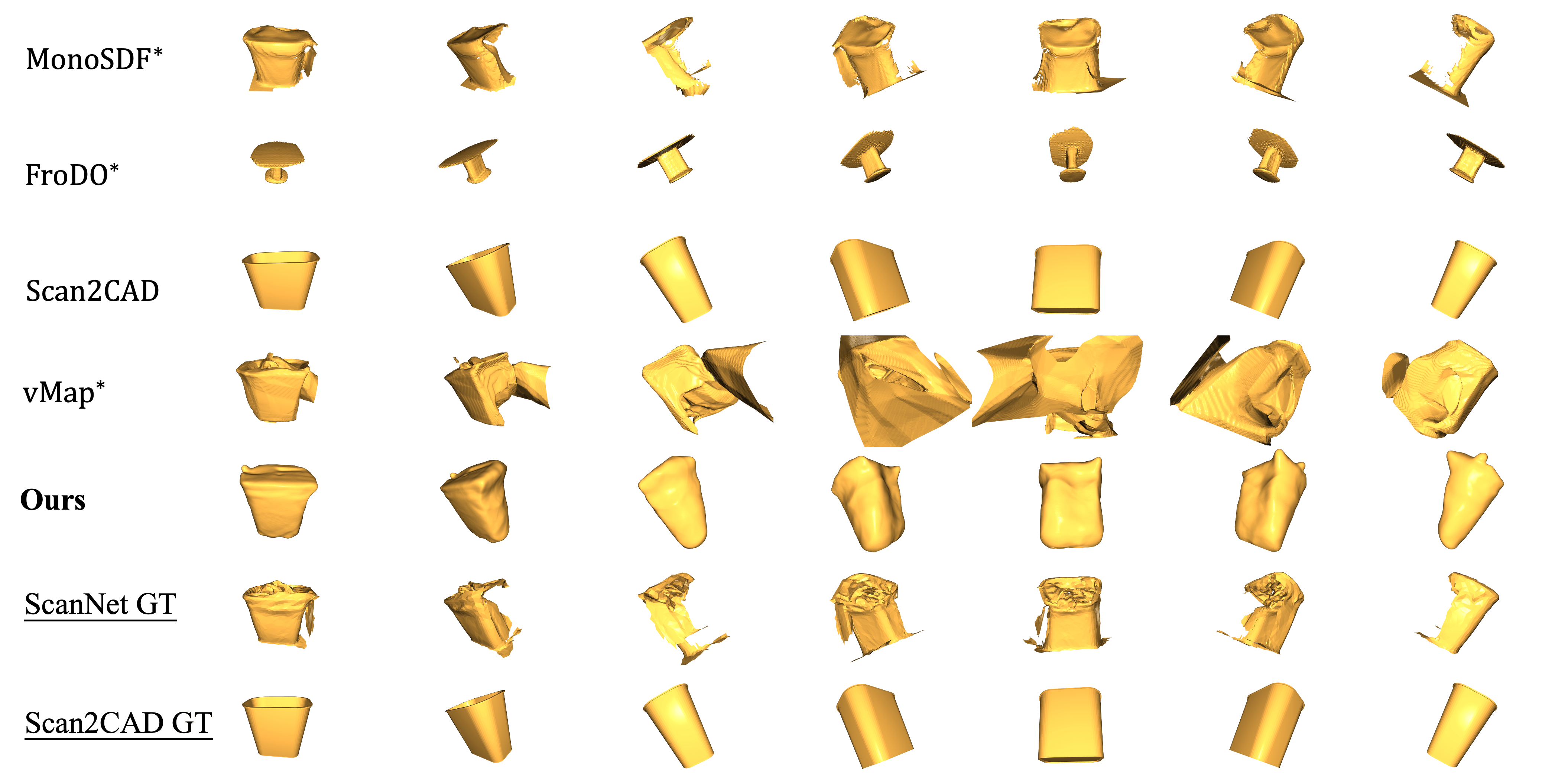}
         \label{fig:vis-obj9}
     \end{subfigure}
     \\[\smallskipamount]
     \begin{subfigure}[b]{1.0\linewidth}
         \centering
         \includegraphics[width=1.0\linewidth]{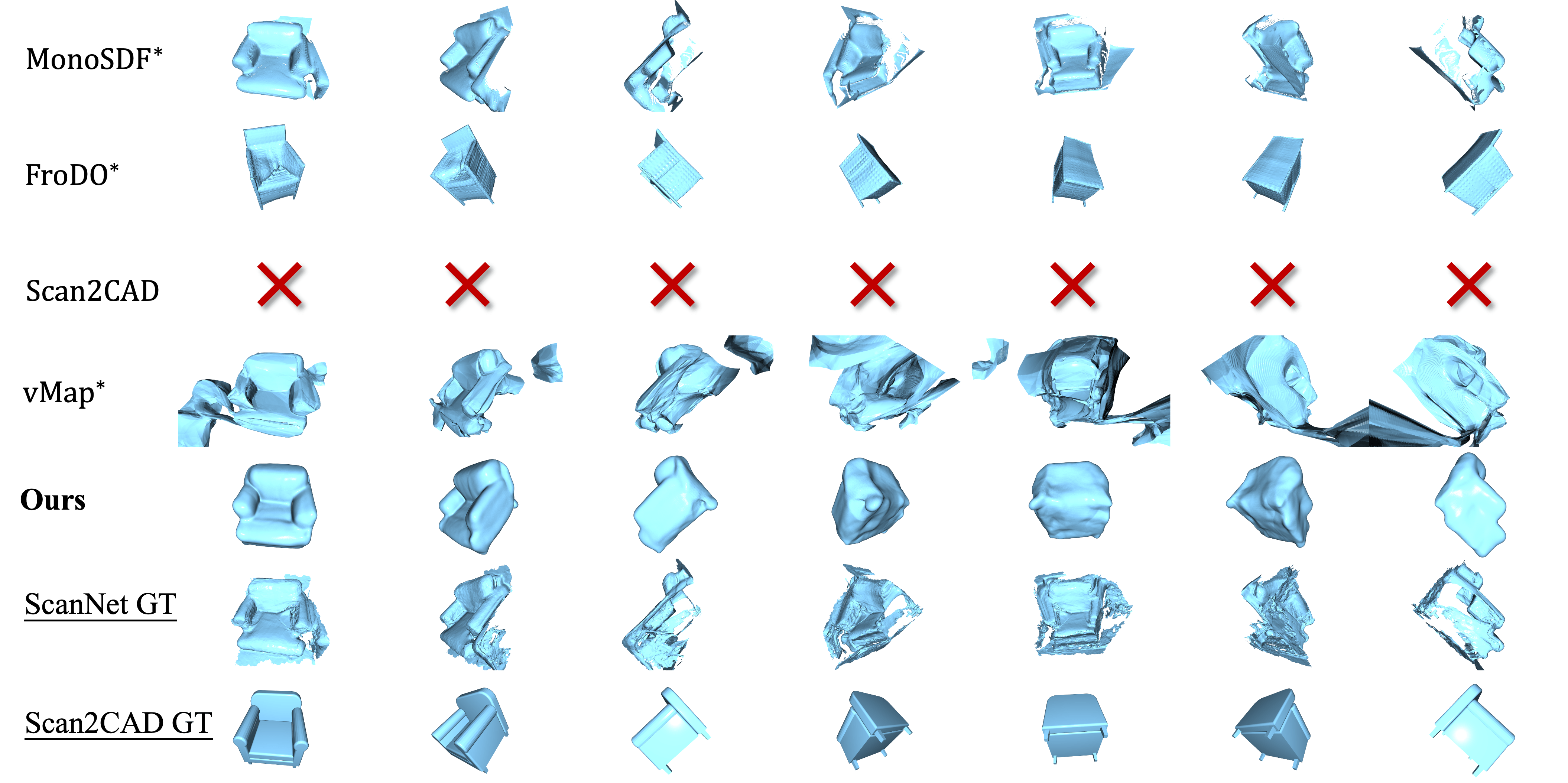}
         \label{fig:vis-obj3}
     \end{subfigure}

\caption{More visualizations of the qualitative comparisons  (5-6).
}
\label{fig:visualization-3}
\end{figure*}

\begin{figure*}[!hbp]
\centering

     \begin{subfigure}[b]{1.0\linewidth}
         \centering
         \includegraphics[width=1.0\linewidth]{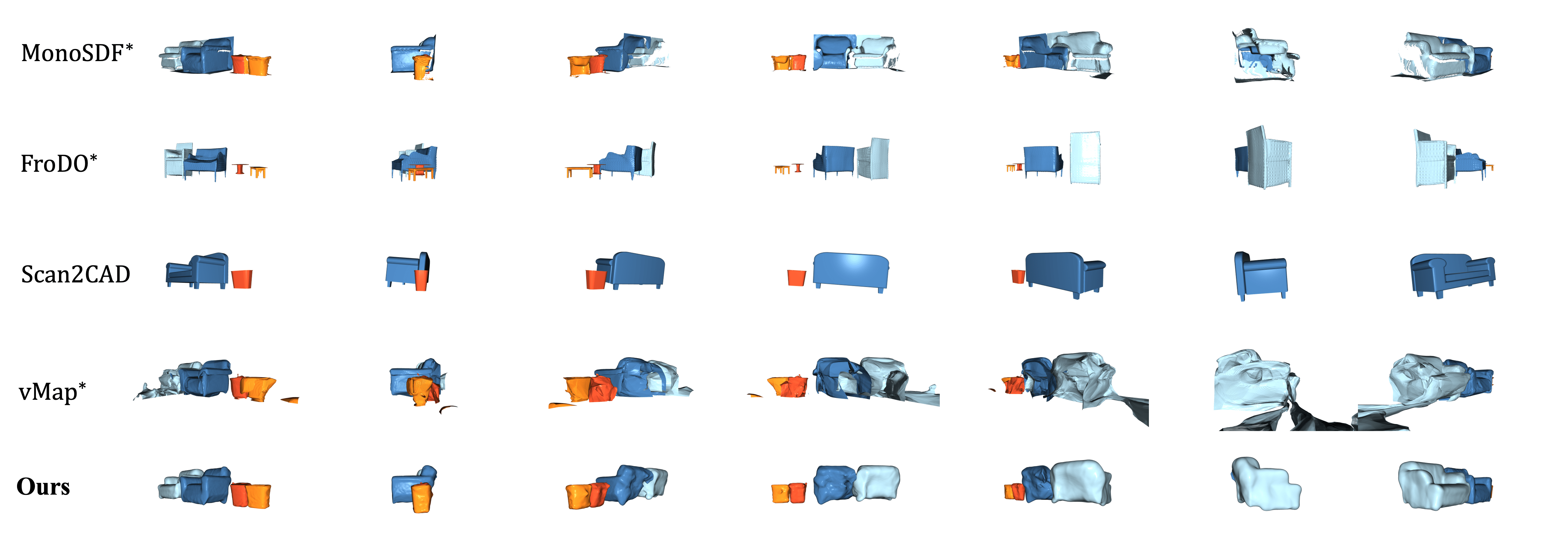}
         \label{fig:vis-scene-orig}
     \end{subfigure}
     \\[\smallskipamount]
     \begin{subfigure}[b]{1.0\linewidth}
         \centering
         \includegraphics[width=1.0\linewidth]{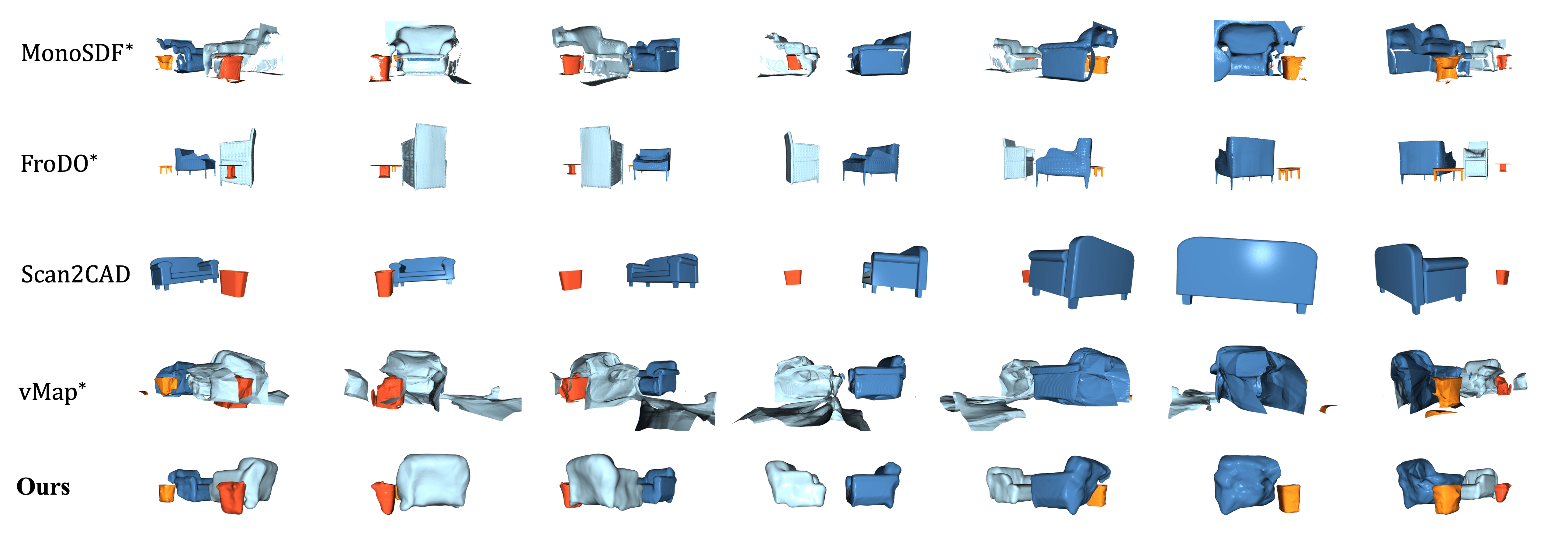}
         \label{fig:vis-scene_trans}
     \end{subfigure}

\caption{More visualizations of the object-level manipulations.
}
\label{fig:visualization-scene}
\end{figure*}

Due to the space limitation, we only show some selected views for each object in the main paper.
In Figure \ref{fig:visualization-1}, \ref{fig:visualization-2}, and \ref{fig:visualization-3}, we compare our method with baseline methods from more perspectives and show the visualization results. 
We also provide the visualizations of object-level manipulation results from more perspectives in Figure \ref{fig:visualization-scene}.
We can observe that O$^2$-Recon produces much better surfaces in the invisible areas of occluded objects.

\end{document}